# 3D-Printed Hydraulic Fluidic Logic Circuitry for Soft Robots

Yuxin Lin,[1] Xinyi Zhou,[1] and Wenhan Cao[1,2] *
[1]School of Information Science and Technology, ShanghaiTech University, Shanghai 201210, China
[2]Shanghai Engineering Research Center of Energy Efficient and Custom AI IC, Shanghai 201210, China
*Correspondence: whcao@shanghaitech.edu.cn

## ABSTRACT

The confluence of soft robotics and fluidic logic have sparked innovations in integrated robots with superior flexibility and potential machine intelligence. However, current fluidically driven soft robots suffer from either a large number of input controlling devices, or limited driving power. Here, we propose a hydraulic fluidic logic circuitry for liquid driven soft robots, leveraging 3D printing technologies. The fundamental building blocks of the system are hydraulic normally-on and normally-off logic gates, namely NOT and AND, along with a multi-connected channel structure functioning as OR. Using minimal-input design principles, the XOR gate can be simplified to only two valves, and used to construct a sensor-free error detector. The design principle can also be extended to full adders, as well as amplifiers, which can greatly improve the flow efficiency of the system. Additionally, taking advantage of the incompressible nature of liquid and optimized logic circuitry using the minimal-input design principle, we present a quadruped soft robot integrated with combinational fluidic logic, realizing bidirectional turtle-like locomotion, controlled by only two inputs. The robot is capable of walking under heavy load and performing controllable underwater locomotion. This hydraulic fluidic soft robotic system utilizes a small number of inputs to control multiple distinct outputs, and alters the internal state of the circuit solely based on external inputs, holding significant promises for the development of microfluidics, fluidic logic, and intricate internal systems of untethered soft robots with machine intelligence.

## INTRODUCTION

Soft robots exhibit extraordinary flexibility and compliance, which enables them to perform diverse functions and tasks in complex environments and daily lives.[1-5] Among the various actuation methods, fluid has been widely utilized in soft robotics as a powering mechanism. Namely, by adjusting the input pressure or volume to regulate the fluidic flow into or out of the soft actuator, one can control its deformation. This has given rise to underwater locomotion mimicking fish,[6-9] pneumatically actuated grippers,[10] soft manipulator arms for object handling,[11, 12] as well as multi-gait soft robots capable of traversing varied terrains[13] and multiple degree-of-freedom soft robots as tools to perform surgeries.[14]

Naturally, as the complexity of tasks performed by soft robot increases, more external inputs are required to control its multi-chamber, and higher degrees-of-freedom. However, additional inputs often require integration of more external devices, which can in turn limit the robot's flexibility and drastically increase the complication. Fortunately, fluidic logic offers a partial solution to this issue.[15] With appropriate valve structures, fluid circuits are allowed to mimic the behavior of digital circuits.[16] For example, multiple NOT gates can be used to generate oscillatory outputs from a single input,[17, 18] enabling periodic posture changes in soft robots.[19-26] A demultiplexer can control multiple actuators with fewer inputs,[27, 28] and combinational logic can be used to control the soft robot, enabling it to change postures under various combinations of input signal waveforms.[29] The involvement of these fluidic logics can greatly reduce the need for external controlling devices such as pumps and solenoid valves, and potentially lead to fully untethered and integrated soft robots with high machine intelligence and computational capabilities.

Similar to its electric equivalents, such as metal-oxide-semiconductor field-effect transistors (MOSFETs),[30] fluidic logic gates can be built to replicate the current-voltage properties of electric transistor circuits in order to execute particular logic functions.[19, 22, 29, 31-36] As an alternative, specialized valve structures can be created in which a single valve serves as a stand-alone logic gate,[37-39] therefore lowering the number of gates needed. With technological advancements, valves for fluidic logic circuitry can be manufactured using soft lithography,[26, 28, 40] 3D printing,[10, 22, 31, 34, 41, 42] sintering [43] and casting molding.[19] The valves can be connected through external piping [33, 44] or fabricated using integrated printing methods to create a stacked structure for the corresponding fluid circuitry.[45]

Although fluidic logic circuits inspired by electronic logic systems have shown considerable potential in controlling soft robots, several issues remain to be addressed. First, systems entirely modeled after MOSFET circuitry require additional pressure sources, either as outputs or as references for standard low pressure.[22, 29, 31, 33] This leads to unnecessary external inputs and an increased number of inner channels. Second, the channels in the circuit are typically considered only as part of a single logic function or a specific logic gate,[28, 34, 46] overlooking their potential for reuse to perform high-complexity logic functions. This leads to wasted spatial resources and increased system complexity.

While minimizing the inputs is required for fluidic logic circuitry, the selection of an appropriate fluid actuation method, being able to transmit sufficient pressure through a compact fluid logic circuitry, is a crucial consideration. Currently, most fluidic logic circuits are pneumatic-based.[21, 22, 29, 33, 34, 38, 39] Pneumatic systems for soft robotics depends on a constant gas supply to maintain high pressure,[47] which makes it challenging in a closed system due to the compressibility of gas. Some untethered soft robots use high-pressure gas tanks as temporary pressure sources,[21, 22] but in such closed systems, the pressure depletes rapidly during locomotion, requiring frequent gas replenishment to sustain the robot's operational state. Furthermore, although traditional microfluidic systems are capable of performing a wide range of logical operations,[37, 48, 49] their low output flow rate results in insufficient actuation power.[38] In contrast, due to the incompressible nature of liquids, hydraulic systems intrinsically offer greater driving forces [50] and output power than the pneumatic ones.[51-53] Additionally, similar to drug delivery, liquids can deliver reactive solutes or chemicals to target locations within the robot.[8, 26, 54] Also, since hydraulic actuation does not cause density change of the robot, it serves as a suitable method for driving soft robots underwater.[55] These factors all emphasize the advantages of using hydraulic actuation in many useful soft robotic applications.

In this work, we propose a design strategy for minimal-input hydraulic logic circuitry. We utilize normally-on and normally-off valves as NOT and AND gates, respectively, and construct OR gates using T-junctions and check valves. These approaches effectively reduce the number of valves required compared to MOSFET-based logic gates. Building upon this, we design a sensor-free error detector based on multiple XOR gates, greatly simplifying the logic circuitry while enabling output waveform verification. In addition to logic operations, we design minimal-input half-adders and full-adders, demonstrating that our approach can be applied to more complex hydraulic logic circuitry. To enhance the efficiency of hydraulic logic circuitry, we develop both active and passive hydraulic flow amplifiers based on the AND gate. Notably, the passive hydraulic flow amplifier achieves output flow rate amplification without additional inputs. We compare the performance of hydraulic and pneumatic actuation, confirming that hydraulic actuation is more effective in driving soft robots. Finally, through combinational logic, we realize bidirectional turtle-like locomotion of a quadruped robot composed of three degree-of-freedom (DoF) actuators, controlled by dual inputs. The robot demonstrates the ability to walk under load and performs controllable underwater walking (Figure 1). Experimental results show that hydraulic fluidic logic circuits effectively control and drive soft robots, offering broad application potential and significant practical value.

## RESULTS

### Hydraulic transistor design and validation

Similar to electric circuitry, fluid flows through channels and controlled by corresponding valves. Since N-channel Metal-Oxide-Semiconductor (NMOS) and P-channel Metal-Oxide-Semiconductor (PMOS) transistors in electrical circuits exhibit similar characteristics, some studies have drawn analogies between normally-on/off valves and PMOS/NMOS transistors to analyze their properties and design Complementary Metal-Oxide-Semiconductor (CMOS) circuits.[23, 31, 33, 56] Here, we design two distinct valve structures exhibiting the same properties. The normally-on and normally-off valve structures, analogous to PMOS and NMOS transistors, are shown in Figure 2A and 2B. The switching of the valves is implemented by the deformation of the membrane, which affects the corresponding flow resistance. The flow resistance distribution in the normally-on and normally-off valves is shown in Figure S1A and S1B.

In the normally-on valve structure, a protrusion is located on the membrane near the upper control chamber. When liquid flows into the control layer, the membrane deforms upward and the protrusion blocks the input port (Figure S1C), causing a sharp increase in the corresponding flow resistance $R_{CPi}$ (Figure S1D) and decrease in the output flow rate $Q_O$. Once the control pressure in the control layer is removed, the membrane moves downward, reopening the valve. The cross-sectional projection of the flow rate distribution when the elastic membrane blocks the input channel is shown in Figure 2C. The flow rate inside the input channel is demonstrated to be nearly 0, indicating that the valve is closed.

In the normally-off valve structure, when control pressure is absent, the valve core moves downward due to input pressure, and the upper gasket blocks the channel, closing the valve. When control pressure is applied, the membrane beneath the valve core is compressed, causing the valve core to lift and open the input channel (Figure 2D). The decrease in the flow resistance $R_{CNi}$ results in the rise of output flow rate $Q_O$. As the input pressure increases, the flow resistance gradually decreases to a certain value, determined by the valve structure, and the flow eventually stabilizes at a constant level (Figure S1E and S1F). The upper surface area of the valve core is designed to be smaller than the lower surface area. This asymmetry in surface areas allows a small pressure applied to the lower control channel to open the upper input channel, effectively enabling a low pressure to control a higher input pressure [37, 39]. The video of the results of the simulation is shown in Video S1.

Based on the simulations, we conduct experiments to obtain curves similar to I-V curves in electric circuitry for the normally-on and normally-off valves (Figure 2E and 2F). The trends of these curves closely align with the simulation results. As the control pressure increases, the normally-on valve closes and the normally-off valve opens, with the switching points shifting as the input pressure changes. These curves closely resemble the I-V characteristics of transistors, demonstrating that the designed normally-on and normally-off valves exhibit PMOS and NMOS behavior.

**Minimal-input logic gate design**

By combining normally-on and normally-off valves based on CMOS, basic logic gates can be achieved (Figure S2A to S2C), and combinations of these logic gates can be used to implement more complex logic capabilities.[34, 39, 48, 57] Usually, achieving basic logic functions in CMOS-based fluid circuits requires combining multiple valves, which needs additional pressure source beyond those required for logic operations. These extra inputs increase the number of redundant channels and add complexity to the overall system.

In hydraulic fluidic logic circuitry, producing an output requires two conditions: the channel must be open, and there must be a corresponding flow source. The opening and closing of the channel are controlled by valves, while the flow source is provided by the inputs. From these, two key principles for designing minimal-input hydraulic logic circuitry can be derived.

First, the minimal number of inputs required for a hydraulic fluidic logic circuit is the sum of the number of inputs needed for its logic function and the number of outputs that are mutually exclusive to all inputs. Second, if an output cannot be synchronized with an input, at least one valve must be introduced to control the switching of the channels when they are interrelated.

Based on these principles, a single normally-on valve can serve as a NOT gate, while a single normally-off valve can function as an AND gate, each performing its corresponding logic function. Additionally, the input signal that is not mutually exclusive to the output signal can act as the flow source, which can be used in the design of OR gate structure (Figure S2D to S2F). This approach significantly reduces the number of valves and inputs required to implement these basic logic functions (Figure S2G to S2I).

In detail, we use a normally-on valve as a NOT gate, with its operating principle and timing diagram shown in Figure 3A. When the control signal is “0” (i.e., no pressure is applied), the output is “1” (i.e., pressure is applied). When the control signal is “1”, the membrane deforms, closing the channel and producing an output of “0”. Based on the first principle for designing minimal-input fluidic logic circuitry, since the input and output of the NOT gate are mutually exclusive, this structure requires a constant pressure source in addition to the control pressure.

For the normally-off valve, the condition for producing an output is the presence of both liquid input and control pressure. The valve’s fluidic behavior aligns with the logic function of an AND gate, allowing the normally-off valve to be directly used as an AND gate (Figure 3B).

For the OR gate, since the output can exist with any of the inputs, no on-off valve is required according to the second designing principle for minimal-input fluidic logic circuitry. Any multi-connected structure can fulfill the OR gate function, and this approach of directly leveraging the structural characteristics of channels or existing materials to manufacture OR gate can effectively reduce spatial complexity and lower costs.[38] In order to control flow direction and prevent backflow, we introduce check valves, ensuring that inputs do not interfere with each other, thereby achieving the function of an OR gate (Figure 3C).

**Design of a sensor-free error detector based on multiple XOR gates**

Based on basic logic operations, more complex logic functions can be decomposed and simplified into several paradigms of cascaded basic logic operations.[29] During logic function execution, certain channels are needed to transmit pressure from external inputs to the corresponding layers of the gates. This transmitted pressure is termed internal sub-input pressure. The relationship between internal sub-input pressure and external input pressure is similar to that between sub-lines and bus lines in CMOS circuitry. As the complexity of the logic function increases, additional external input pressures are introduced, may resulting in significantly increased number of logic gates. Consequently, internal sub-input pressures rise substantially, and the added channels further complicate the circuitry structure.

By reusing different channels to fully utilize their logical functions, complex logic operations can be achieved through simple valve combinations with minimal inputs. For example, the XOR function can be constructed using only two NOT gates and one OR gate (Figure 4A).

In the proposed XOR gate, the methods for reusing channels are as follows: the control signal from one NOT gate is used as the pressure source for the output of the other NOT gate. By connecting the outputs of both NOT gates through an OR gate and utilizing the relationship among valve output, flow source, and channel opening or closing, the XOR logic operation can be achieved (Figure 4B). Compared to the original XOR logic function expression, this approach eliminates the need for the AND operation, significantly simplifying the fluidic logic circuitry and reducing the number of inputs.

For the XOR operation, which generates an output when the two inputs differ, multiple XOR gates can be used to compare the consistency between input signals and an anticipated standard reference signal $S_E$. When an inconsistency is detected, an output is generated on $S_E$, indicating that the corresponding output does not meet expectations, which may lead to errors such as $E_1$ or $E_2$ (Figure 4C and 4D). This comparison method is highly significant in the field of fluidic logic,[58] as it enables verification of the correctness of fluidic logic system without the need for sensors.

By reusing input channels as flow sources and fully utilizing the structural characteristics of the valves, the number of external inputs required for the entire XOR circuitry and the internal sub-inputs for the valves are significantly reduced compared to previous designs and methods (Figure 4E). Notably, previous designs inspired by CMOS circuits often require one or two additional inputs to function as "Voltage Drain (VDD)" or "Ground (GND)" when performing logic functions, leading to unnecessary external input pressure. Although the atmosphere can serve as a constant pressure source, channels directly connected to the atmosphere are unsuitable for underwater environments. For closed system, the "0" state in the valve chamber can only be achieved by connecting to a standard low-pressure source generated by additional channels. Furthermore, the previous methods for forming complex logic circuitry were merely connecting and combining basic logic gates, which could increase the internal sub-input pressure required for each gate and result in the need for more channels and valves. Consequently, this increases the overall complexity of the system compared to this work.

The physical diagram of the sensor-free error detector circuitry is shown in Figure 4F. The introduction of check valves not only ensures that the OR gate functions properly but also prevent the reference signal from mixing with other output channels, as demonstrated in Video S2 using different colored dyes.

**Design of hydraulic full adder logic unit**

In addition to performing logic operations, hydraulic logic circuitry can also execute arithmetic operations by combining logic gates. In modern computers, the adder is a fundamental component within Arithmetic Logic Unit (ALU) for binary arithmetic. So, constructing an adder system using simple circuits and fewer inputs is of significant importance in hydraulic logic circuitry.

The configuration of a half-adder is shown in Figure 5A. By reusing the input signals as flow sources for the outputs and fully utilizing the structural characteristics of the valves, we simplify the hydraulic logic circuitry for producing the half-adder's "Sum" output into an XOR gate consisting of two NOT gates and one OR gate. Meanwhile, Input 2 serves as the flow source for the half-adder's "Carry" output. The experimental setup is shown in Figure 5B.

When both inputs of the half-adder system are in the "0" state, there is no flow source, and thus, no output. When either Input 1 or Input 2 is applied individually, the "Sum" output is produced due to the properties of the XOR gate, but since the AND gate condition is not satisfied, the "Carry" output remains 0. When both Input 1 and Input 2 are applied, the XOR gate closes, and the AND gate channel opens to produce an output (Figure 5C). The timing diagram generated by the hydraulic logic circuitry is shown in Figure 5D and Video S3, which satisfies the half-adder truth table while effectively reducing the number of required valves and inputs.

Building upon the minimal-input hydraulic half-adder circuitry, we connect two half-adders and one AND gate to achieve the hydraulic full-adder circuitry (Figure 6A). The truth table is shown in Figure 6B. When the logic gates are cascaded as shown in Figure 6C, the valves successfully perform the full-adder function (Figure 6D and Video S4), demonstrating that the minimal-input design principle is equally effective when extended to multi-stage hydraulic fluidic logic systems.

**Design of active and passive flow amplifiers**

As the structure of hydraulic logic circuitry becomes more complex, the flow resistance in the circuit substantially increases due to the added valves and longer pipelines, which reduces the output flow velocity and ultimately degrades the operation performance.[15, 16, 59] These issues are particularly significant when driving large-volume soft actuators, as they require a high enough flow rate for effective actuation. [60] Therefore, it is crucial to achieve the required flow rate amplification without compromising the accuracy of the output results.

Interestingly, the flow amplification effect can be achieved through an AND gate. We observe that when one input of the AND gate is held constant at the "1" state, its output is entirely determined by the other input (Figure 7A). Since this valve can use a smaller control pressure input to regulate a larger pressure input, we conduct experiments to measure the relationship between the output pressure $P_o$ and the control pressure $P_c$ under a larger input pressure $P_s$ (Figure 7B). The results show that $P_o$ closely follows the waveform of $P_c$, and they remain largely consistent within a certain range. This occurs because once the channel is opened, $P_c$ and $P_o$ act together on the upper and lower surfaces of the membrane. If there is a significant difference between $P_c$ and $P_o$, the membrane deforms, causing the channel to either open or close, establishing a new mechanical equilibrium.

When the channel is open, the output flow rate $Q_o$ equals the sum of the control flow $Q_c$ and the source flow $Q_s$, with the introduction of additional flow providing a corresponding gain $A_i$ (Figure 7C). This gain can be further amplified by connecting multiple amplifiers in parallel. In our experiments, we measure the relationship between the output flow rate $Q_o$ and the number of connected amplifiers under the condition of $P_s$ = 60 kPa and $P_c$ = 35.6 kPa (Figure 7D). The results confirm that the amplifier effectively increases the flow rate, and the gain rises as the number of parallel flow amplifiers increases. Using three active flow rate amplifiers under these input pressures provided a 30% flow gain, and the increase in the number of passive flow amplifiers can cause a corresponding steady increase in the flow rate when the number of passive flow amplifiers is larger than $N(N \geq 2)$. By comparing the parallel system of multiple amplifiers to a multi-resistance parallel system, we explain the mechanism of gain generation and increasing, as shown in Figure S3A.

With amplified output flow rate, the fluidic logic system can now be used to actuate soft actuators. We apply this amplification system to simultaneously drive three soft actuators, observing the gain effect on the bending velocity of the actuator using three parallel active flow rate amplifiers (Figure 7E and Video S5). Under otherwise identical conditions, the time required for the amplification system to achieve a 70° bend is 81.9% of the time required for that without an amplifier design. The reduced flow gain, compared to Figure 7D, is due to the fact that as the actuator bends, the internal pressure gradually approaches the input pressure, leading to the decreasing in the flow rate. The decreasing pressure difference reduces the flow amplification factor.

Since the active hydraulic flow amplifier requires an additional input as a flow source, it adds a burden to the system by increasing the volume of input pipelines. To address this, we modify the amplifier by using the original hydraulic logic circuitry's input as the source of additional flow. In this setup, the amplifier functions are equivalent to the original hydraulic logic circuitry but with a parallel connection to a flow resistance, thereby reducing the overall resistance of the hydraulic circuit (Figure S3B). This modification enables flow rate amplification under the same input conditions (Figure 7F).

To find out the performance of this passive hydraulic flow amplifier, we use a multiple XOR gates detector. Two syringes of the same specification are connected to the detector to compare the volume of output from two channels over the same period of time. The syringe connected to $E_1$ is labeled $C_A$, and the one connected to $E_2$ is labeled as $C_{NA}$. Based on the results (Figure 7G and Video S6), the volume of liquid in $C_A$ is greater than that in $C_{NA}$ over the same time period, indicating that the output with amplifier gain is larger than the output without gain, with a gain effect of 32.2%. The experiment demonstrates that the passive hydraulic flow amplifier still provides significant flow amplification without additional input sources.

**Hydraulic logic circuitry for soft robot actuation and controlling**

A significant advantage of applying hydraulic logic circuitry to control soft robots lies in its ability to utilize combinational logic to independently control multiple outputs with fewer inputs. This enables the actuation of 3-DoF soft robots, allowing for various postures. We design the hydraulic logic circuitry shown in Figure 8A to control a quadruped soft robot, consisting of four 3-DoF legs. This circuit design is based on the half-adder circuitry from Figure 5C. By removing the OR gate, the hydraulic circuitry is able to use two inputs to control three outputs independently (Figure 8B). Without the restriction of the one-way valve in the OR gate, fluid at the output can return to the syringe through the input channel (Figure 8C). A check valve is introduced to address the issue of slow backflow caused by narrow channel $R_{C2}$ (Figure S4A and S4B). With good airtightness of Stereolithography (SLA) printed actuators, the soft robot and its integrated stacking hydraulic logic circuitry can form a closed system and be applied across various scenarios.

This closed system can achieve high-pressure actuation of the soft robot through energy exchange with external device, without the need to replenish the fluid medium—highlighting a key advantage of hydraulic systems over pneumatic systems. The robot's movement mimics the crawling motion of a turtle, with clockwise or counterclockwise limb movements propelling the body forward or backward. This efficient control strategy enables versatile locomotion with minimal input complexity.

The process of moving rightward is illustrated in Figure 8D. When only $P_{i1}$ is applied, the $C_2$ chambers of the four actuators expand, lifting the soft robot. Once $P_{i1}$ is set to “0” and only $P_{i2}$ is active, the $C_1$ chambers expand, driving the body to the right. Meanwhile, with $P_{i1}$ at zero, the elastic potential energy stored in the deformed $C_2$ chambers is released, causing a rebound effect. Under the influence of internal pressure $P_{o2}$ generated by this rebound, fluid flows back from the higher-pressure $C_2$ chambers to the input channels. Subsequently, both $P_{i1}$ and $P_{i2}$ are set to “0”, and the liquid stored in the $C_1$ chambers is forced back to the input side under rebound pressure, returning the actuators to their initial undeformed state. By alternating between these three input states, the soft actuators’ ends exhibit a clockwise rotational movement, which continuously propels the quadruped soft robot to the right (Figure S4C and Video S7).

The pressure changes in the input and output channels during this process are shown in Figure 8E and Video S8. These variations highlight how the control of internal pressure enables the coordinated motion of the soft robot, achieving efficient locomotion through the sequential actuation of different chambers.

For the leftward locomotion, the soft actuators’ ends exhibit a counterclockwise rotational movement (Figure 8F). Since the $C_3$ chambers are controlled by the output of an AND gate, they expand only when both $P_{i1}$ and $P_{i2}$ are applied. The backflow from these chambers occurs when $P_{i1}$ is “0” and $P_{i2}$ is “1”, resulting in the waveform and speed of leftward locomotion that differ from those of rightward locomotion (Figure 8G, Figure S4D and Video S8).

Building on the success of using dual-input hydraulic logic circuitry to control the quadruped soft robot’s bidirectional locomotion, we further explore its ability to perform various specific tasks. Figure 8H demonstrates the robot’s walking capability while carrying a load. We test two load weights, 200 g and 250 g, and found that the soft robot could walk normally with both, demonstrating its ability to carry more than three times its own weight (Video S9).

Additionally, when the soft robot is applied to underwater scenes, the hydraulic logic circuitry can maintain stable actuation and allow the robot to walk underwater without causing uncontrolled buoyancy changes, which pneumatic soft robots usually suffer due to density variations during actuator expansion (Figure 8I and Video S10). The system’s excellent airtightness ensures there is no internal fluid leakage and prevents external fluid infiltration. We also find that in a closed system with a fixed fluid volume, hydraulic actuation can deliver higher pressure (Figure S5A to S5E) and achieve superior actuation performance compared to pneumatic actuation (Figure S5F to S5I). As a result, beyond terrestrial environments, the hydraulic soft robot shows strong potential for applications in high-pressure underwater environments comparing to pneumatic soft robot.

## DISCUSSION

In this work, distinct from previous pneumatic logic circuitry, we use liquid as the pressure transmission medium and hydraulic actuation to form a closed system that integrates the logic circuitry within the soft robot. This system can transmit high pressure without requiring additional fluid supplementation, offering higher energy transmission efficiency compared to pneumatic systems. We propose minimal-input design principles and apply them to simplify fluidic logic circuitry. These approaches enable the design and fabrication of minimal-input versions of NOT, AND, and OR gates. Building on the designed NOT, AND, and OR gates, we streamline the XOR gate by reusing input signals between valves and design a sensor-free error detector consisting of multiple XOR gates. Additionally, we implement arithmetic hydraulic logic functions, such as a half-adder and a full-adder. Compared to previous work, our design significantly reduces the number of valves and external inputs, and simplify the inner channel structure required for the logic circuitry. The minimum-input adder functions as a black-box model, directly performing operations on all inputs to produce the result without requiring additional inputs in specific parts of the fluid logic circuit to assist the calculating process. This approach more closely aligns with the characteristics of a computer. It is worth noting that our approaches do not rely on specific valves structures and are applicable to most fluid logic circuitry composed of normally-on and normally-off valves. As fluidic logic system complexity increases, our methods for reducing input numbers and simplifying system structure hold tremendous potential value.

Furthermore, based on the structural characteristics and logic function of the AND gate, we design and fabricate both active and passive flow amplifiers. By incorporating these flow amplifiers, we effectively reduce the overall flow resistance of the system, achieving substantial flow amplification with a single valve, thereby improving the efficiency of the output drive. This illustrates that the analogy between circuits and fluid circuitry can be applied not only to structural design but also to performance optimization of fluid circuitry. Similarly, because this flow amplification primarily relies on the logical function of the AND gate and the method of reducing equivalent flow resistance by through parallel resistance, it can also be applied to AND gates designed and fabricated through other methods.

For hydraulic logic control and actuation of soft robots, we use a turtle-like gait quadruped soft robot as the experimental platform to demonstrate minimal-input hydraulic logic circuitry-driven control. The valves and channels are designed in a stacked configuration, and SLA printing is used to manufacture the valve structure and soft actuators. Compared to Fused Deposition Modeling (FDM) printed soft robots that require intricate design to enhance sealing [10, 22], the structure printed in this work exhibits superior airtightness. A secondary curing process is employed to connect the components, forming a closed hydraulic system that includes the pressure source, hydraulic logic circuitry, and soft actuators. The enhanced airtightness ensures more reliable operation in high-pressure environments, particularly for hydraulically-driven systems. In this closed system, with an input pressure of 80 kPa, the liquid is alternately pushed into the actuators and expelled by actuator rebound, forming an internal circulation system that drives the soft robot's controllable bidirectional locomotion.

Hydraulic logic circuitry offers the potential to implement sophisticated control schemes, enabling complex motions in robots driven by simple inputs. Notably, the entire control process is purely fluidically executed, eliminating the need for external electronic circuits. The system also features an expandable design, allowing interconnected units to form more intricate circuitry. The key innovation lies in the minimal-input hydraulic logic circuitry, which uses as few inputs and valves as possible, achieving diverse functionality with reduced structural complexity. Furthermore, its closed-system allows deployment in various environments. This novel characteristic provides a promising design paradigm for the internal systems of untethered soft robots, offering enhanced versatility and maneuverability through fluidic logic.

## METHODS / EXPERIMENTAL PROCEDURES

### Finite element analysis of valve flow characteristics

The finite element analysis simulations are conducted using COMSOL Multiphysics® 6.2. The Fluid-Structure Interaction module is used to simulate the flow characteristics of the normally-on/off valve. The valve model is created in SOLIDWORKS and imported into COMSOL for simulation. The detailed process of simulation and results are provided in Supplementary Information.

### Basic hydraulic logic gates fabrication

Based on the materials used and the different structures of the valves, we develop distinct fabrication schemes for the three valve designs. The OR gate is constructed by connecting a T-shaped junction and two check valves, which are commonly used components for hose connections. The use of silicone tubing for the connections resulted in minimal loss along the pipeline due to its short length. For the NOT gate and the AND gate, we employed stereolithography 3D printing technology to manufacture the shells and the right part of the AND gate membrane. The thinnest wall thickness of the shell is 0.5 mm. The NOT and AND gates for the half-adder and full-adder are printed using a Liquid Crystal Display (LCD) curing printer (HALOT-MAGE S, Creality Inc., China) with transparent LCD curable resin (3D printing UV-sensitive resin, Qiefeng Inc., China). The transparent logic gates of the hydraulic circuitry for soft robot actuation are printed with a high-speed masked SLA 3D printer (Form 4, Formlabs Inc., USA) using transparent SLA curable resin (Clear Resin V5, Formlabs Inc., USA). The elastic membrane is fabricated by pouring and molding a mixture of silicone (Ecoflex 00-30, Smooth-On, Inc., USA) part A and part B in a 1:1 ratio.

We use a secondary curing method to assemble the valve. The specific procedure involves applying uncured resin to the contact surfaces of the two printed parts, pressing them together, and then placing them in a post-curing machine (Form Cure, Formlabs Inc., the USA) at 60°C for 15 minutes. This process fully cures the resin and ensures the material achieves optimal mechanical properties.

### Basic hydraulic logic gates fabrication

Based on the materials used and the different structures of the valves, we develop distinct fabrication schemes for the three valve designs. The OR gate is constructed by connecting a T-shaped junction and two check valves, which are commonly used components for hose connections. The use of silicone tubing for the connections resulted in minimal loss along the pipeline due to its short length. For the NOT gate and the AND gate, we employed stereolithography 3D printing technology to manufacture the shells and the right part of the AND gate membrane. The thinnest wall thickness of the shell is 0.5 mm. The NOT and AND gates for the half-adder and full-adder are printed using a Liquid Crystal Display (LCD) curing printer (HALOT-MAGE S, Creality Inc., China) with transparent LCD curable resin (3D printing UV-sensitive resin, Qiefeng Inc., China). The transparent logic gates of the hydraulic circuitry for soft robot actuation are printed with a high-speed masked SLA 3D printer (Form 4, Formlabs Inc., USA) using transparent SLA curable resin (Clear Resin V5, Formlabs Inc., USA). The elastic membrane is fabricated by pouring and molding a mixture of silicone (Ecoflex 00-30, Smooth-On, Inc., USA) part A and part B in a 1:1 ratio.

We use a secondary curing method to assemble the valve. The specific procedure involves applying uncured resin to the contact surfaces of the two printed parts, pressing them together, and then placing them in a post-curing machine (Form Cure, Formlabs Inc., the USA) at 60°C for 15 minutes. This process fully cures the resin and ensures the material achieves optimal mechanical properties.

### Flow characteristic testing of normally-on and normally-off valves

We conduct tests on the flow characteristics of the normally-on and normally-off valves to verify their actual flow performance. The tests utilize pressure sensors (XGZP6847A, 0-500 kPa, CFSensor, China) and a flow meter (YRS-SFR, 1-200 ml/min, Shanghai Urisi Instrument Co., China). Both the sensors and the flow meter are connected to the valves using silicone tubing. We calibrate the flow meter using a syringe pump (LSP04-1A, Longer Pump, China). When the pump flow rate is set to 20 ml/min, the flow meter output an average of 2.28 pulse signal changes per second, corresponding to a flow rate of 0.147 ml/s per pulse. The sensors and flow meter are connected to an Arduino board (Arduino UNO, ARDUINO, Italy), with the sampling frequency set to 4 Hz. We use two small dispensing machines (BXH-982A, Ben Xin He, China) to provide stable source pressure and control pressure. For a given source pressure, the output flow rate is measured by increasing the control pressure to the same fixed value over five trials. The wiring diagrams for the normally-on and normally-off tests are shown in Figure S6 and Figure S7.

### Experiment of hydraulic logic circuitry functional testing

We use three small dispensing machines as stable pressure sources for the experiment. Although the system utilizes a pneumatic dispenser, it functions solely as a pressure input source for the hydraulic logic circuitry, with only energy transfer occurring between the two systems. The output channels are connected to the corresponding dispensing syringes or actuators. For this procedure, we use pressure sensors (XGZP6847A, 0-500 kPa, CFSensor, China), which are connected to an Arduino board, with the sampling frequency set to 5 Hz.

### Experiment of comparison between hydraulic and pneumatic actuation

We design a series of experiments to compare the driving effects of hydraulic and pneumatic systems when the input fluid volumes are the same. We use a compression testing machine (Universal Testing System, Instron Inc., the USA) to evaluate the load-bearing capacity of hydraulic and pneumatic actuators under static pressure conditions (Figure S5A and S5B). The results are shown in Figure S5C. Due to the incompressibility of liquids, hydraulic actuators are capable of bearing greater loads compared to pneumatic actuators. To compare the supporting effects of the two driving methods, we design a structure as shown in Figure S5D, employing four actuators to support a square platform. The platform compresses a pressure gauge fixture (FORCE GAUGE, Yueqing Handpi Instruments Co., Ltd. China) to assess the driving forces generated by hydraulic and pneumatic actuation when injecting equal volumes of fluid. As shown in Figure S5E, hydraulic actuation provides a more significant driving force when injecting the same volume of fluid, which aligns with the characteristic of high energy transfer efficiency in hydraulic systems. We inject the same volume of liquid and gas into the soft actuator using a syringe pump to compare the effects of both actuation methods on the bending angle of the actuator (Figure S5F).

We compare the variation in the bending angle of the actuator with different initial volumes as the injected volume changed. The results, shown in Figure S5G, indicate that when the input fluid volumes are equal, hydraulic actuation outperforms pneumatic actuation. Furthermore, the effectiveness of pneumatic actuation is also influenced by the initial volume. The impact on the actuator's bending angle directly affects the driving control of the soft robot. We compare the lifting performance of a quadruped soft robot with the same fluid volume input. As shown in Figure S5H and S5I, when 6.8 ml of liquid and gas are injected, hydraulic actuation lift the soft robot's body three times farther than pneumatic actuation. Based on these results, we conclude that in a closed environment without the need for additional input supplements, hydraulic actuation demonstrates a significantly superior performance compared to pneumatic actuation when using the same volume.

### Experiment of hydraulic logic circuitry for quadruped soft robot actuation and control

In the closed system of a soft robot composed of hydraulic logic circuitry and soft actuators, two pressure sources are required for actuation. Dispensing machines are selected as the primary pressure source; however, it can be replaced by alternative pressure sources capable of generating equivalent pressure effects, such as peristaltic pumps, syringe pumps, or hydraulic cylinders. The actuators used in the soft robot are manufactured using SLA-cured elastic resin (Elastic 50A resin V2, Formlabs Inc., USA), while the body is constructed from transparent resin. The system for monitoring the soft robot's motion states and corresponding input and output pressures is shown in Figure 8D to Figure 8G. The three output channels connected to one of the actuators can be interfaced with three corresponding sensors to collect hydraulic pressure data from the respective chambers (Figure S8).

**Data availability**
The authors declare that the data supporting the findings of this study are available within the paper and its supplementary information files.

## ACKNOWLEDGMENTS

Use this section to acknowledge contributions from non-authors and list funding sources. Because this section contains important information and many funding bodies require inclusion of grant numbers here, please check it carefully.

## AUTHOR CONTRIBUTIONS

W. C. conceived and supervised the project. Y. L. and X. Z. conducted the fabrication of soft actuators. Y. L. and W. C. designed the logic gate system. Y. L. and W.C. drafted the initial manuscript. All authors participated in the discussion of the results, analysis of experimental data, and revision of the manuscript.

## DECLARATION OF INTERESTS

The authors declare no competing interests.

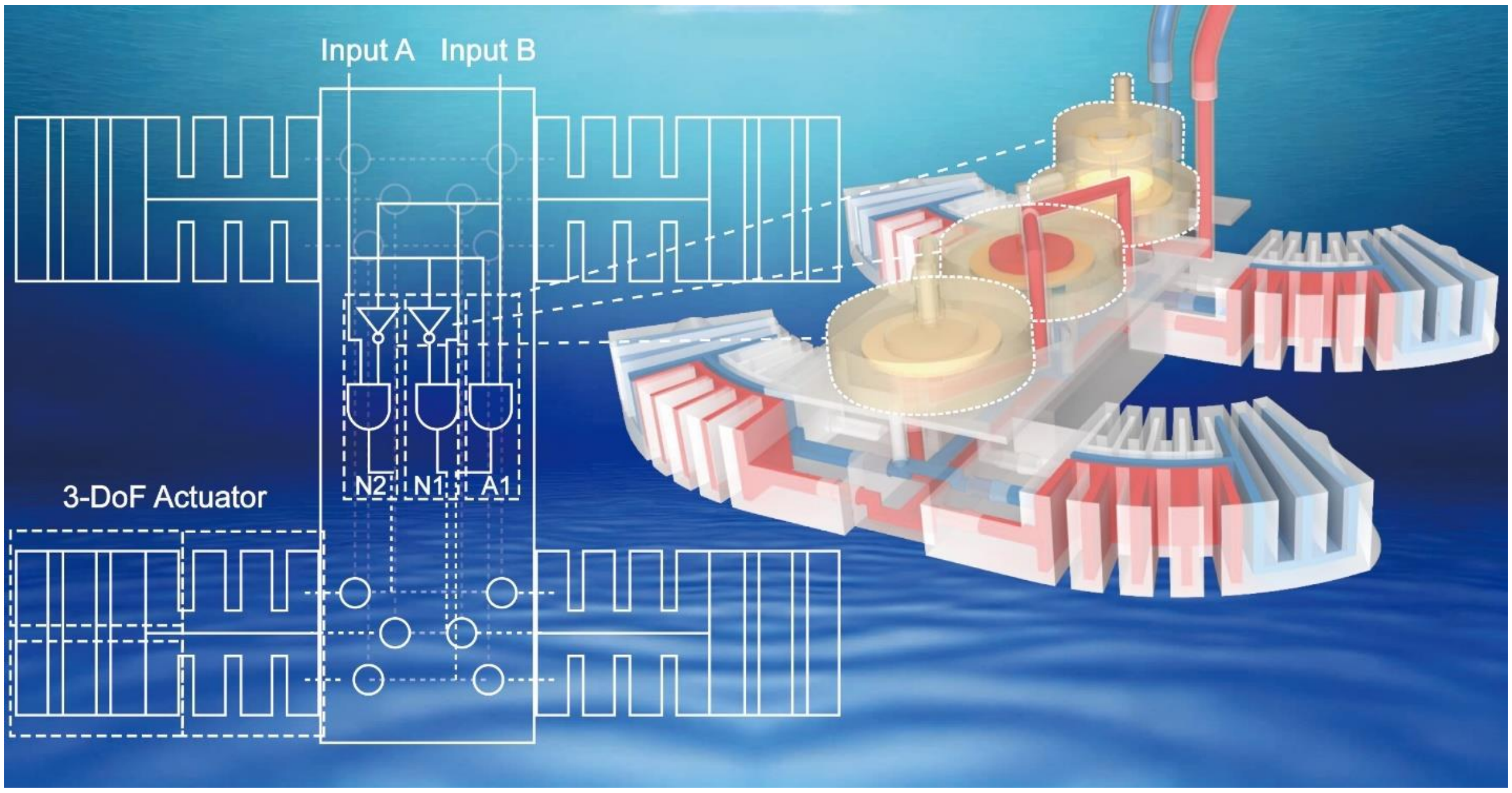

Figure. 1 Overview of hydraulic logic circuitry controlling a quadruped soft robot for underwater operations.

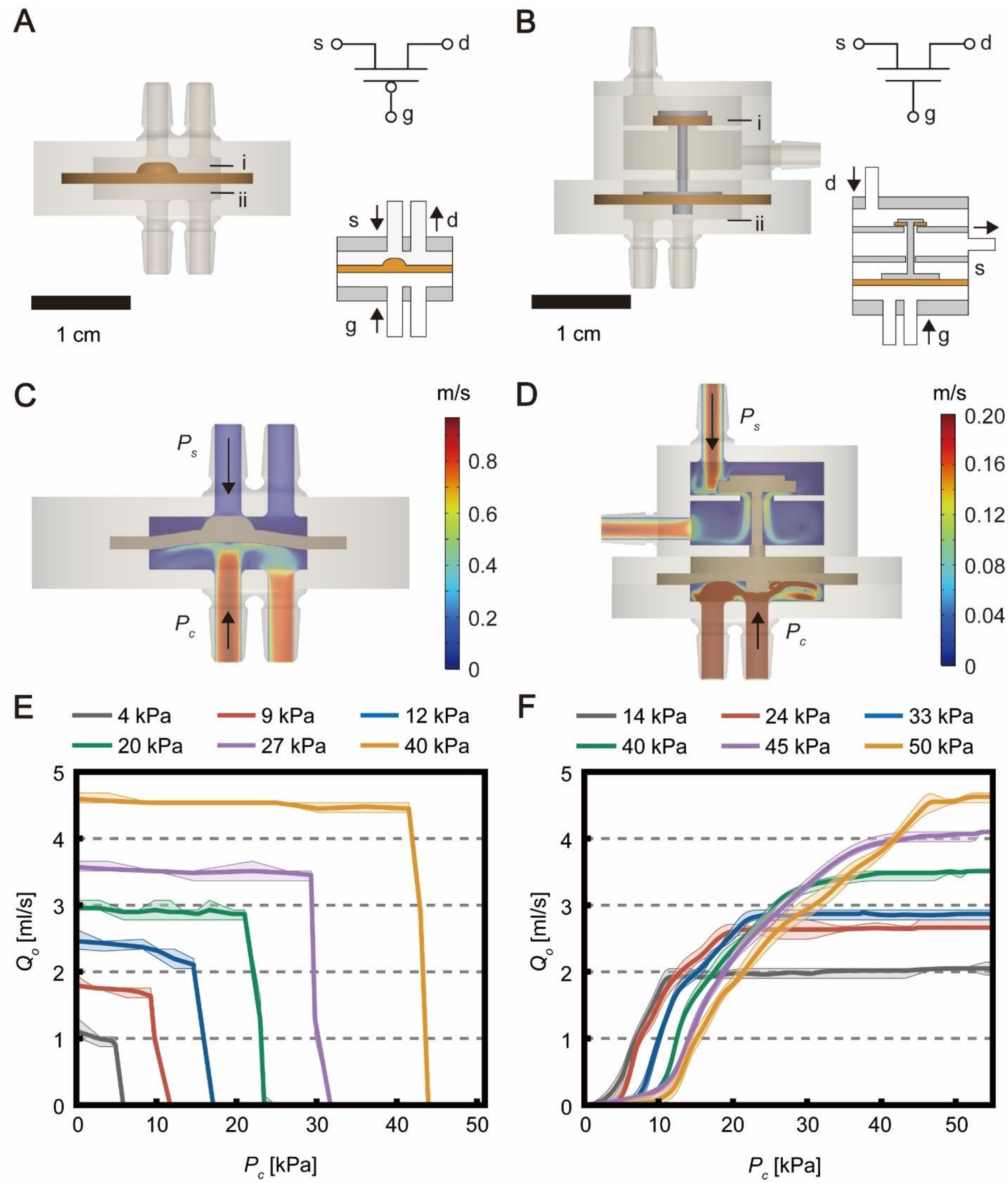

Figure. 2 Design and functional validation of normally-on/off valves. (A) Model of the normally-on valve. (B) Model of the normally-off valve. Each valve consists of an input layer (i) and a control layer (ii). The two channels of the input layer serve as nodes s and d, while the control layer serves as node g for PMOS and NMOS, respectively. (C) The cross-sectional projection of the flow rate distribution of the normally-on valve when the channel is closed under an input pressure $P_i = 100$ Pa and control pressure $P_c = 500$ Pa. (D) Flow rate distribution of the normally-off valve when the channel is open under an input pressure $P_i = 10$ Pa and a control pressure $P_c = 500$ Pa. (E) The output flow rate $Q_o$ as a function of control pressure $P_c$ for the normally-on valve. (F) The output flow rate $Q_o$ as a function of control pressure $P_c$ for the normally-off valve.

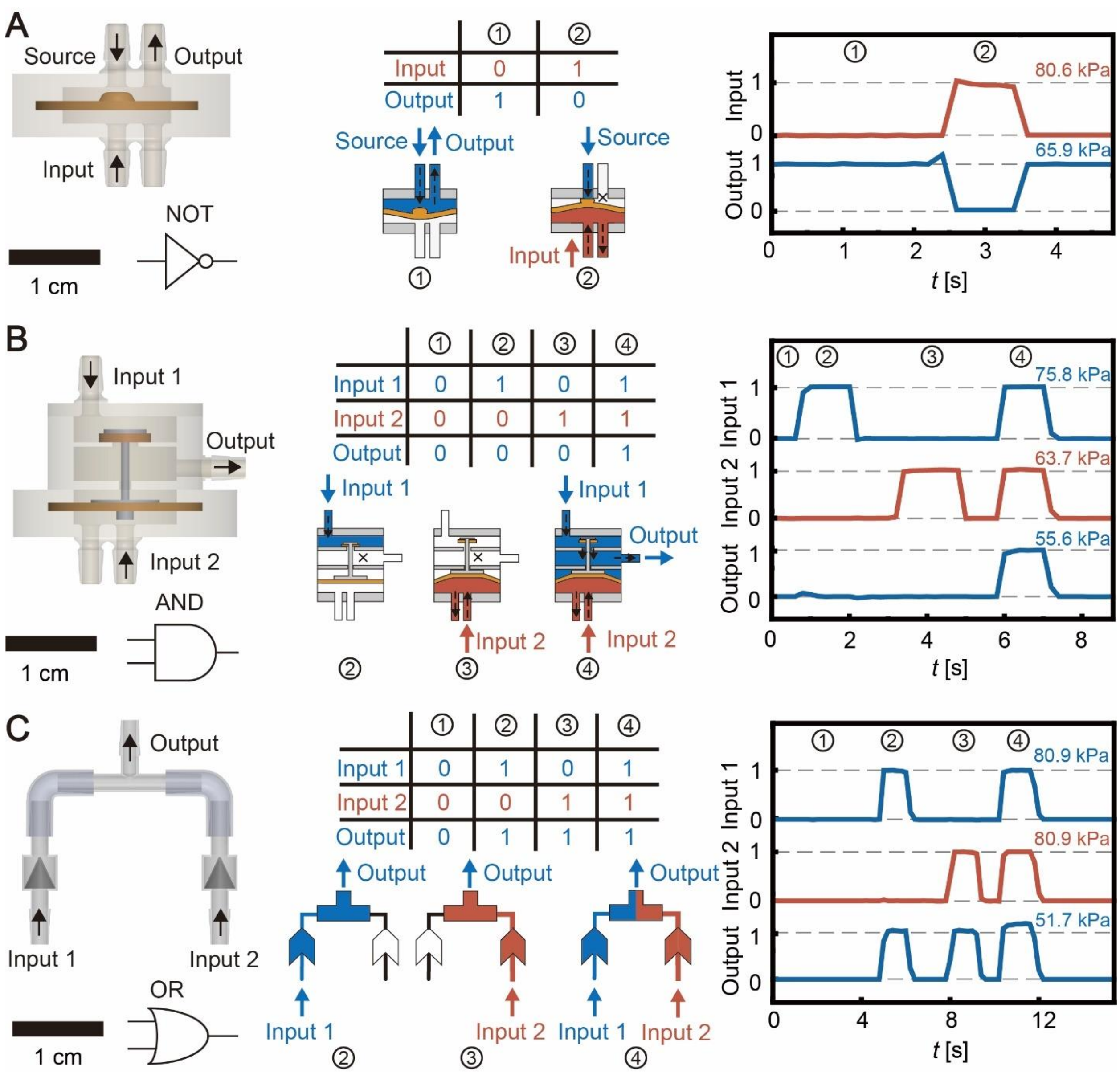

Figure. 3 Minimal-input logic gate design and operating principle. (A) Operating principle and timing diagram for using a normally-on valve as a NOT gate. (B) Operating principle and timing diagram for using a normally-off valve as an AND gate. This method fully utilizes the valve inputs without requiring additional source pressure, thereby reducing the number of required inputs. (C) Operating principle and timing diagram for using multichannel connectors and check valves as an OR gate. This method does not require a diaphragm valve and can prevent crosstalk between fluidic channels, effectively achieving the corresponding logical operation.

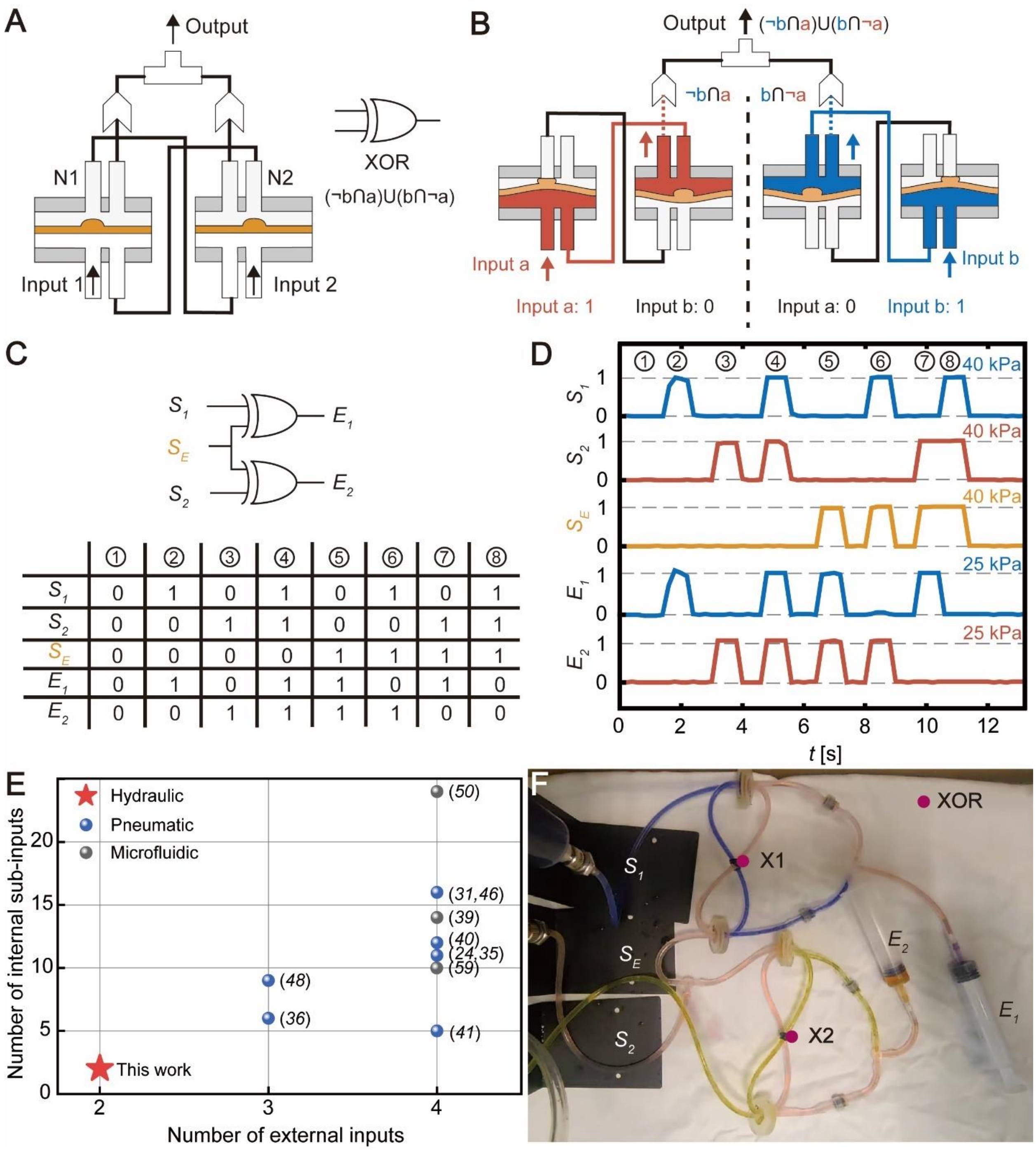

Figure. 4 The simplified design method of the XOR gate and the use of multiple XOR gates as a sensor-free error detector. (A) Schematic of an XOR gate composed of two NOT gates and one OR gate. (B) The operating principle and simplified design method for the XOR gate. (C) The use of multiple XOR gates as a detector. Signal $S_E$ serves as the expected standard reference signal. When other input signals deviate from $S_E$, the corresponding XOR gate generates an output, indicating that an error has occurred. (D) Experimental results of the sensor-free error detector based on multiple XOR gates. The timing diagram aligns with the truth table of the corresponding system. (E) Comparison of the number of inputs and gates between methods from previous work and this work. (F) A physical diagram of the sensor-free error detector composed of multiple XOR gates.

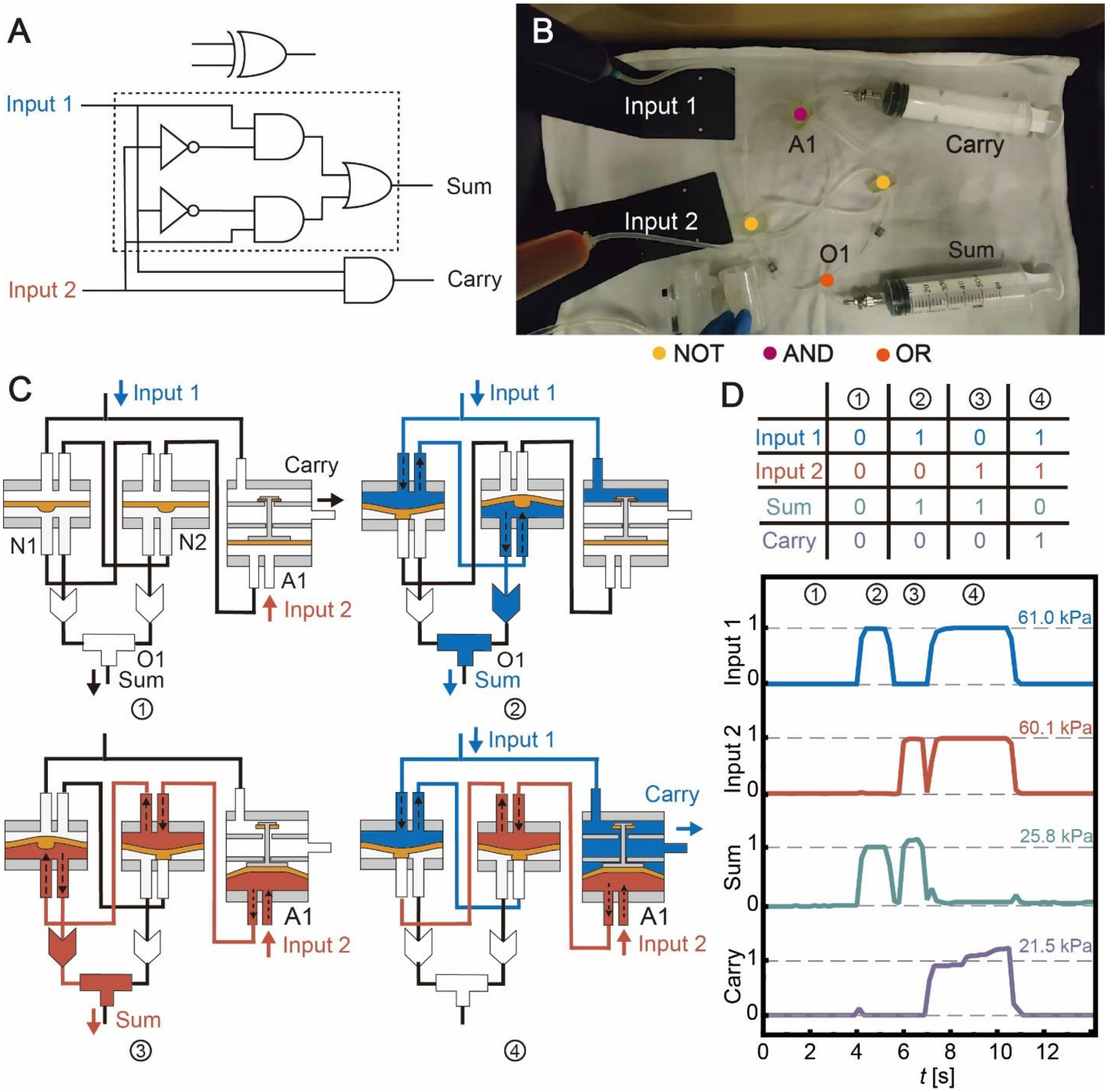

| | ① | ② | ③ | ④ |
|---|---|---|---|---|
| Input 1 | 0 | 1 | 0 | 1 |
| Input 2 | 0 | 0 | 1 | 1 |
| Sum | 0 | 1 | 1 | 0 |
| Carry | 0 | 0 | 0 | 1 |

Figure. 5 Minimal-input hydraulic half-adder design. (A) The composition of a half-adder using basic logic gates. The XOR logic gate is composed of the logic gates within the box. (B) A physical diagram of the hydraulic half-adder circuitry. (C) The operating principle of the hydraulic half-adder circuitry. (D) The truth table of the half-adder and the timing diagram of the hydraulic half-adder circuitry.

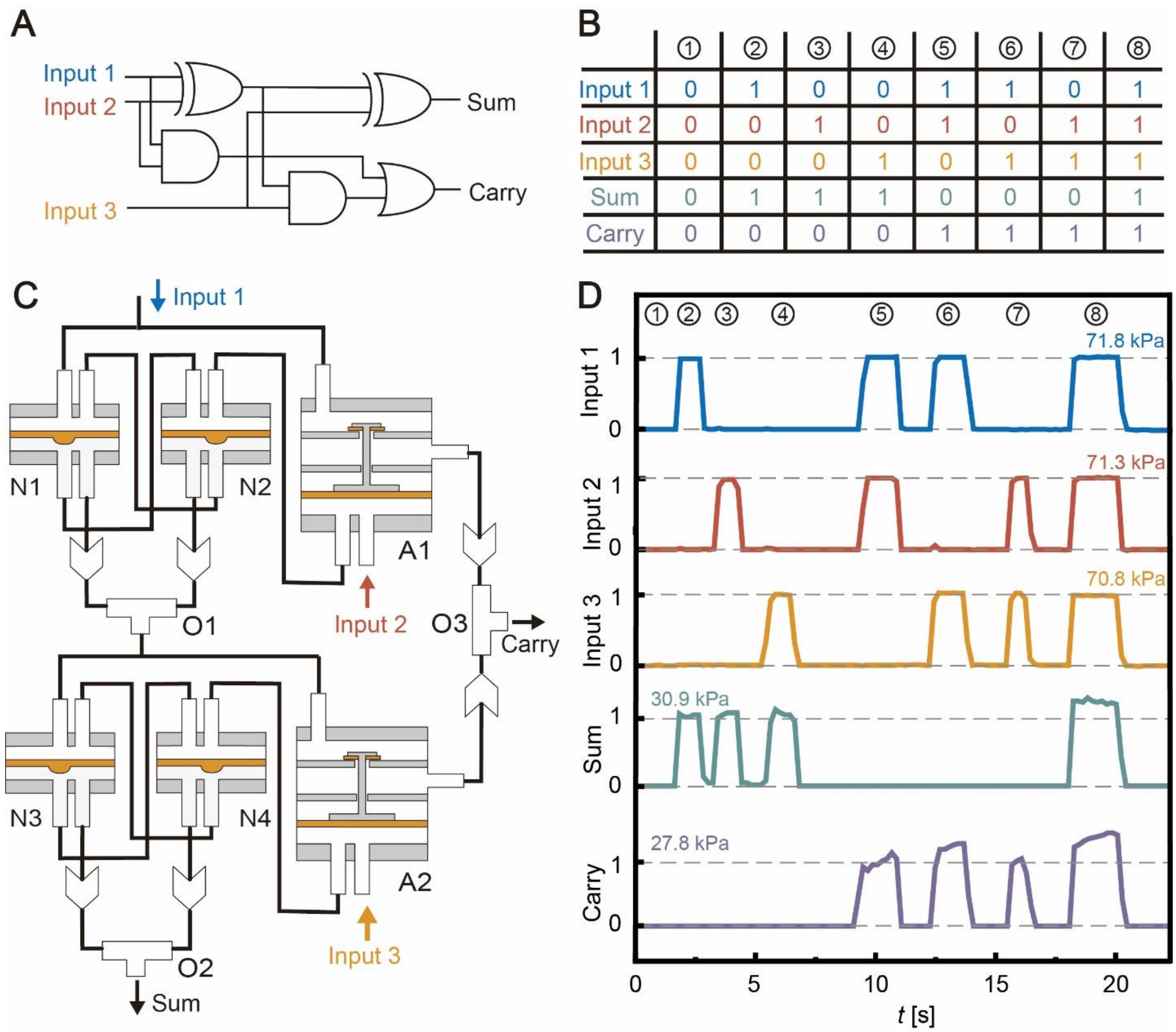

| | ① | ② | ③ | ④ | ⑤ | ⑥ | ⑦ | ⑧ |
|---|---|---|---|---|---|---|---|---|
| Input 1 | 0 | 1 | 0 | 0 | 1 | 1 | 0 | 1 |
| Input 2 | 0 | 0 | 1 | 0 | 1 | 0 | 1 | 1 |
| Input 3 | 0 | 0 | 0 | 1 | 0 | 1 | 1 | 1 |
| Sum | 0 | 1 | 1 | 1 | 0 | 0 | 0 | 1 |
| Carry | 0 | 0 | 0 | 0 | 1 | 1 | 1 | 1 |

Figure. 6 Minimal-input hydraulic full-adder design. (A) The composition of a full-adder using logic gates. A full-adder can be viewed as a cascade of two half-adders and an AND gate. (B) The truth table of the full-adder. (C) The schematic of the hydraulic full-adder circuitry. (D) The timing diagram of the hydraulic full-adder circuitry.

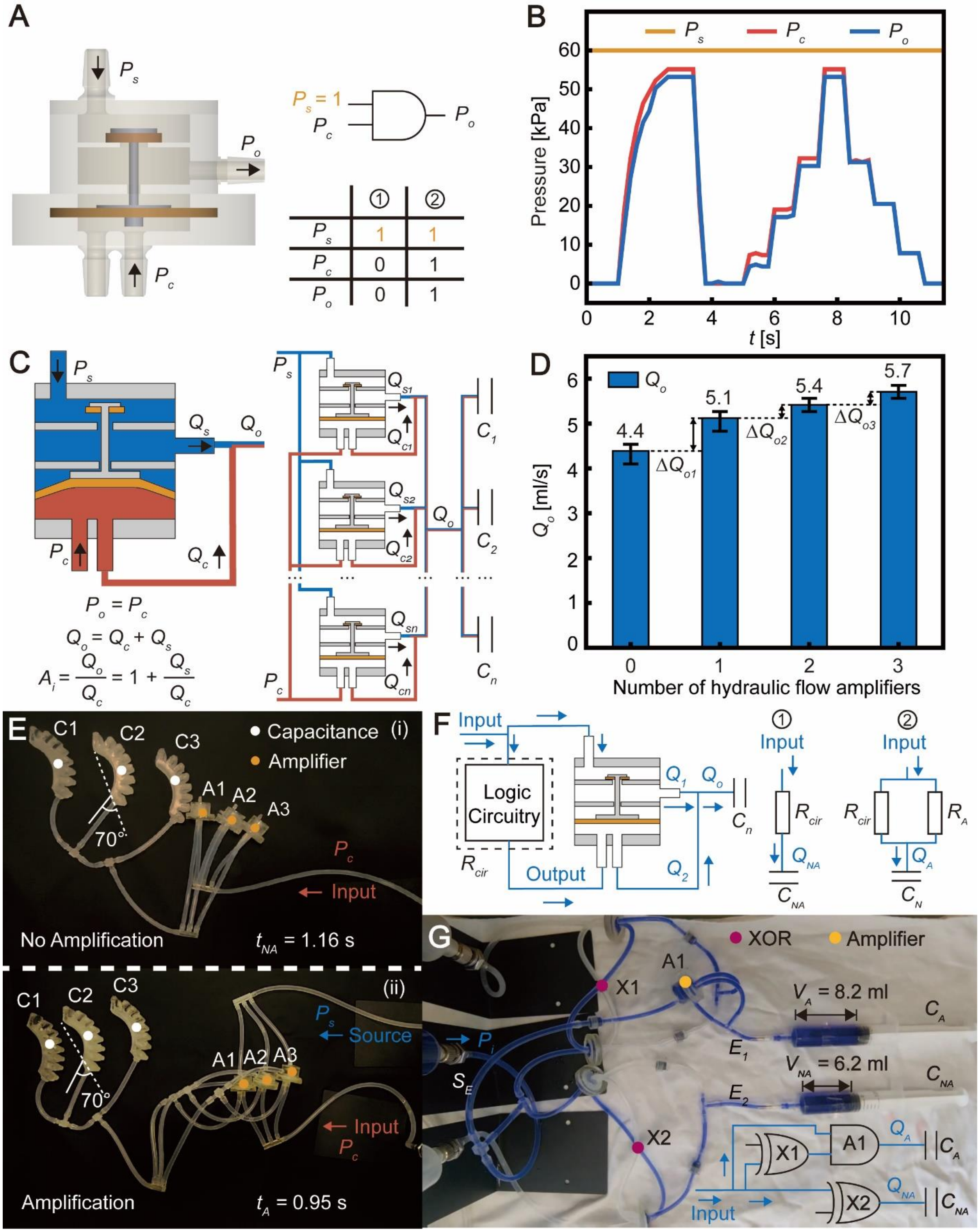

Figure. 7 Design and functional verification of active/passive hydraulic flow amplifier. (A) Truth table of AND gate when one of the inputs is fixed to "1" state. (B) The relationship between output pressure $P_o$ and control pressure $P_c$ under a higher source pressure $P_s$ for AND gate. (C) Design and operating principle of active hydraulic flow amplifier. AND gate produces an output flow $Q_o$ that is derived from the original flow $Q_c$ and another flow $Q_s$ provided by higher source pressure $P_s$. The additional input of flow brings gain, and the gain can be amplified by paralleling multiple amplifiers. (D) The relationship of output flow-rate and the number of hydraulic flow amplifiers when $\boldsymbol{P_s = 60\ kPa}$ and $\boldsymbol{P_c = 35.6\ kPa}$. (E) The comparison of driving effect between no amplification hydraulic circuitry (i) and amplification hydraulic circuitry (ii). Three actuators serve as the flow capacitance and three active amplifiers are used. The source pressure $\boldsymbol{P_s = 60\ kPa}$ and control pressure $\boldsymbol{P_c = 35.6\ kPa}$. (F) Design and operating principle of passive hydraulic flow amplifier. In this situation input for the hydraulic logic circuitry serves as the source pressure for the amplifier. (G) The gain effect of a passive flow amplifier on the flow under input pressure $\boldsymbol{P_i = 35.6\ kPa}$. This amplifier provides gain for one output in the multiple XOR detector. The amplification effect of the passive flow amplifier is determined by comparing the liquid volume in two syringes over the same time period.

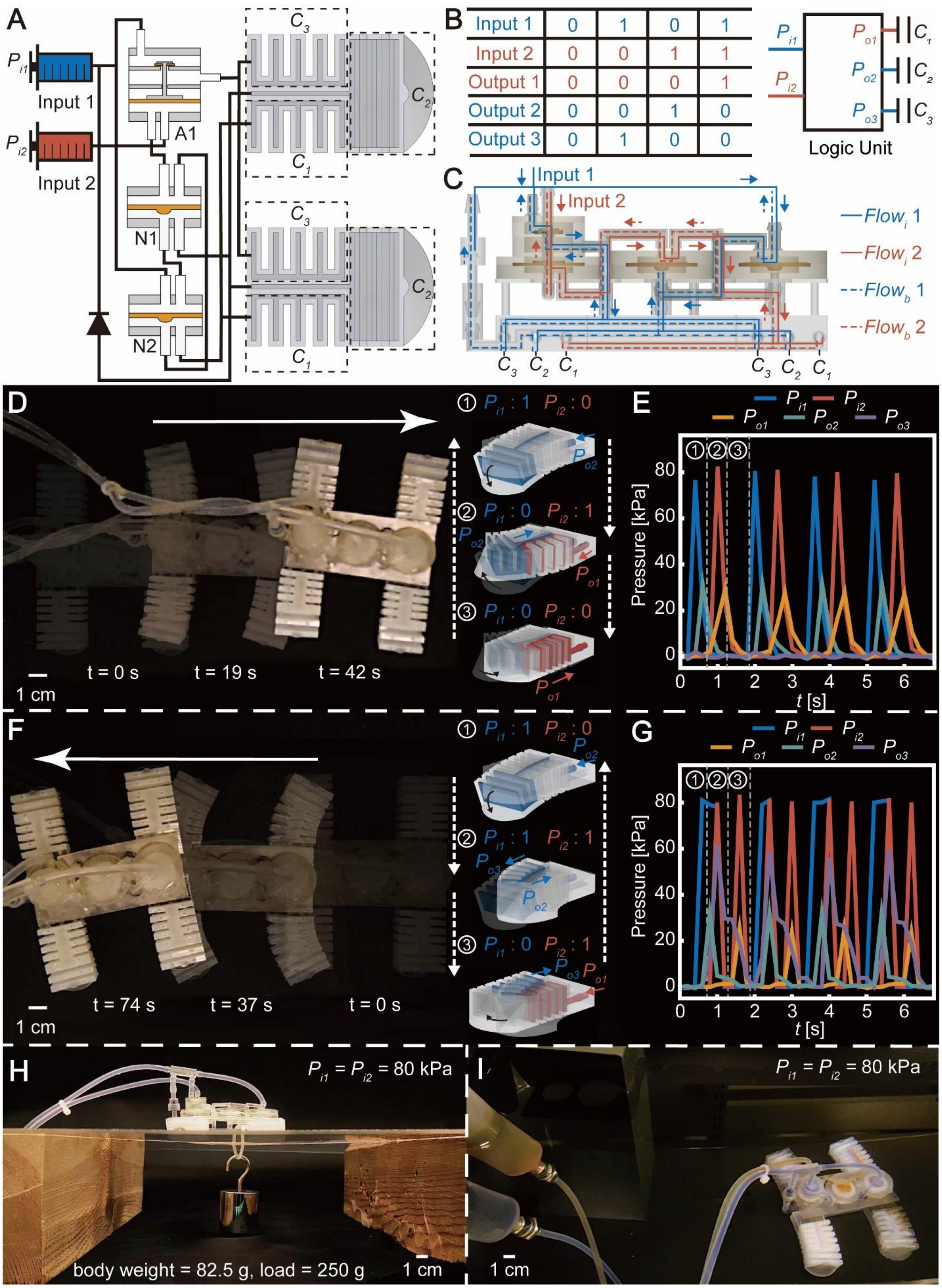

A
$P_{i1}$
Input 1
$P_{i2}$
Input 2
A1
N1
N2
$C_3$
$C_2$
$C_1$
B
Input 1 0 1 0 1
Input 2 0 0 1 1
Output 1 0 0 0 1
Output 2 0 0 1 0
Output 3 0 1 0 0
$P_{i1}$
$P_{i2}$
$P_{o1}$
$P_{o2}$
$P_{o3}$
Logic Unit
C
Input 1
Input 2
$Flow_i$ 1
$Flow_i$ 2
$Flow_b$ 1
$Flow_b$ 2
D
① $P_{i1}$ : 1 $P_{i2}$ : 0
② $P_{i1}$ : 0 $P_{i2}$ : 1
③ $P_{i1}$ : 0 $P_{i2}$ : 0
t = 0 s
t = 19 s
t = 42 s
1 cm
E
Pressure [kPa]
$t$ [s]
F
① $P_{i1}$ : 1 $P_{i2}$ : 0
② $P_{i1}$ : 1 $P_{i2}$ : 1
③ $P_{i1}$ : 0 $P_{i2}$ : 1
t = 74 s
t = 37 s
t = 0 s
1 cm
G
Pressure [kPa]
$t$ [s]
H
$P_{i1}$ = $P_{i2}$ = 80 kPa
body weight = 82.5 g, load = 250 g
1 cm
I
$P_{i1}$ = $P_{i2}$ = 80 kPa
1 cm

Figure. 8 Quadruped soft robot controlled by closed-loop hydraulic logic circuitry. (A) Schematic of using closed hydraulic logic circuitry to control soft robot. The hydraulic logic circuitry generates a driving effect without the need for constant replenishment of the pressure transmission medium from external sources, with only energy exchange occurring. (B) Truth table of the hydraulic logic circuitry. By manipulating the two input pressures, $P_{i1}$ and $P_{i2}$, three independent outputs $P_{o1}$, $P_{o2}$, $P_{o3}$ can be generated, which can be utilized to independently control the three chambers $C_1$, $C_2$, $C_3$ of the 3-DoF actuator. (C) Cycle routes for different inputs for this hydraulic logic circuitry. $Flow_i$ denotes that the liquid injected from the syringe into the wall through the corresponding channels under external pressure. $Flow_b$ denotes that the liquid stored in the chamber that is returned to the syringe under rebound pressure after the external pressure is removed. (D) With the cycle alternating input states of ① $P_{i1}: 1\ P_{i2}: 0$, ② $P_{i1}: 0\ P_{i2}: 1$, ③ $P_{i1}: 0\ P_{i2}: 0$, the end of the 3-DoF actuator rotates clockwise, resulting in a rightward movement of the soft robot. (E) The variation of input pressures $P_{i1}$, $P_{i2}$ and output pressures $P_{o1}$, $P_{o2}$, $P_{o3}$ over time during rightward movement. (F) With the cycle alternating input states of ① $P_{i1}: 1\ P_{i2}: 0$, ② $P_{i1}: 1\ P_{i2}: 1$, ③ $P_{i1}: 0\ P_{i2}: 1$, the end of the 3-DoF actuator rotates counterclockwise, resulting in a leftward movement of the soft robot. (G) The variation of input pressures $P_{i1}$, $P_{i2}$ and output pressures $P_{o1}$, $P_{o2}$, $P_{o3}$ over time during leftward movement. (H) Load-carrying mobility of quadruped soft robots under input pressures $P_{i1} = P_{i2} = 80$ kPa. (I) Underwater locomotion for quadruped soft robot under input pressure $P_{i1} = P_{i2} = 80$ kPa.

**SUPPLEMENTARY INFORMATION**

**1. Supplemental simulation procedures and results.**

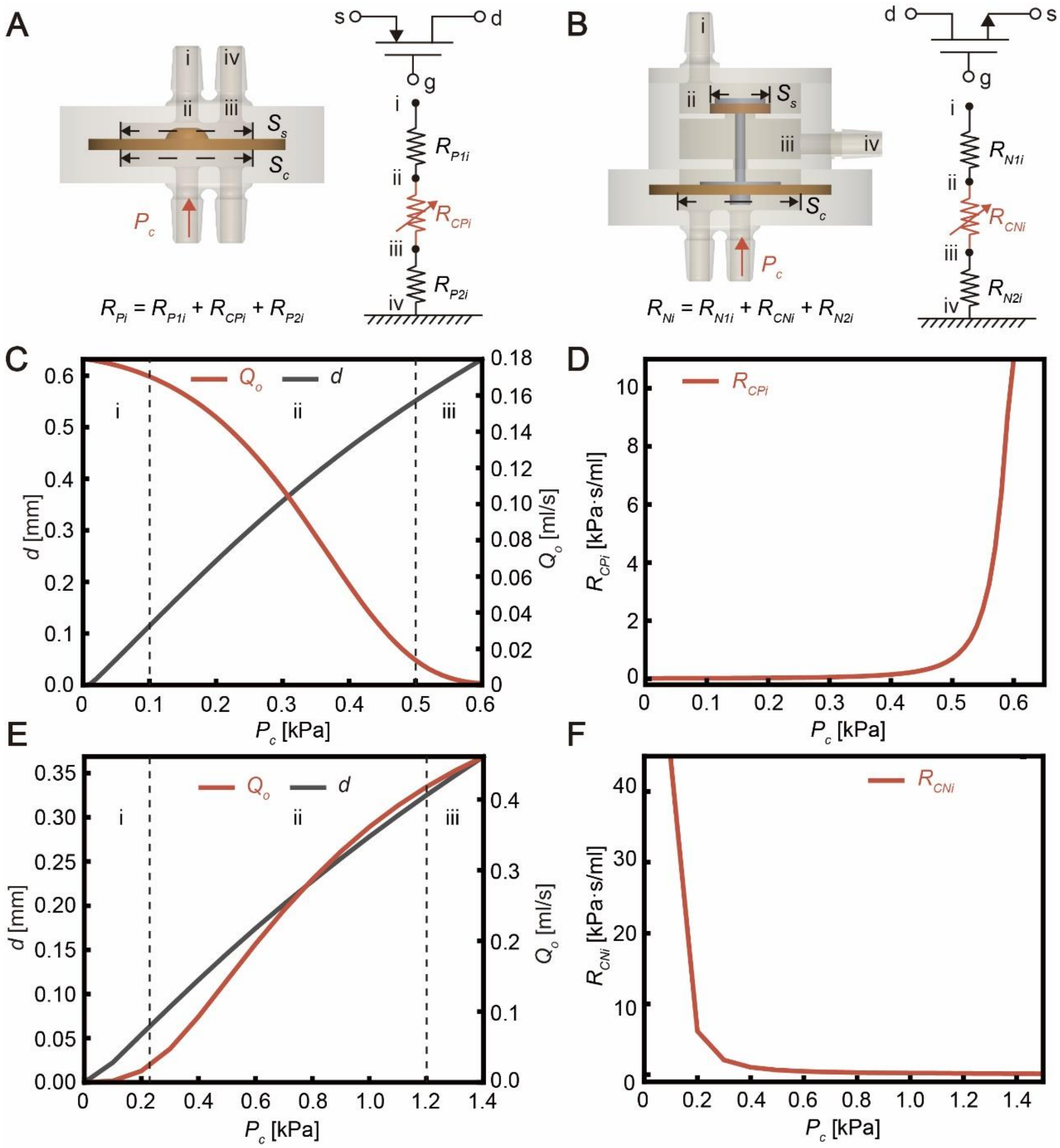

**Figure S1. Simulation results of normally-on/off valves.** (A) Flow resistance analysis of input layer of normally-on valve. (B) Flow resistance analysis of input layer of normally-off valve. $S_s$ and $S_c$ represent the areas of the membrane affected by the source pressure and control pressure, respectively. Flow resistance of input layer consist of resistance in channels $R_{1Pi}$, $R_{2Pi}$, $R_{1Ni}$, $R_{2Ni}$ and resistance controlled by membrane $R_{CPi}$, $R_{CNi}$ for PMOS and NMOS, respectively. (C) Simulation results of normally-on valve. The increasing of control pressure $P_c$ results in the displacement of the top of membrane $d$ and the decrease of the flow rate $Q_o$. Regions (i), (ii), and (iii) correspond to the MOSFET Shockley first-order model's cutoff region (i), linear region (ii), and saturation region (iii), respectively. (D) The relationship between flow resistance $R_{CPi}$ and control pressure $P_c$. The rapid rise in resistance means the channel is gradually closing. Regions (i), (ii), and (iii) correspond to the MOSFET Shockley first-order model's saturation region (i), linear region (ii), and cutoff region (iii).(E) Simulation results of normally-on valve. The increasing of control pressure $P_c$ results in the displacement of the top of membrane $d$ and the rise of the flow rate $Q_o$. (F) The relationship between flow resistance $R_{CNi}$ and control pressure $P_C$. The rapid drop in resistance means that the channel goes from being completely closed to gradually opening up.

In COMSOL, we obtain the “input layer” and “control layer” fluid domains for the normally-on and normally-off valves through the “Domain Separation” operation. These domains are set as deformable. The material for the fluid domain is defined as “H2O (water) [liquid]”, with the liquid density and dynamic viscosity set to the software’s default values. The flow is defined as laminar, and compressibility is set to “incompressible flow”. For the material properties of the solid in the valve structure, the density is set to 1190 kg/m$^3$, the Young's modulus to 3.2×10$^9$ Pa, and the Poisson’s ratio to 0.35. For the elastic membrane material, the density is set to 970 kg/m$^3$, the Young’s modulus to 1×10$^5$ Pa, and the Poisson’s ratio to 0.49. The boundary between the solid material and the fluid domain is defined using the “Fluid-Structure Interaction” boundary.

We add input and output channels for both the "control layer" and "input layer", where the source pressure for the "input layer" is defined as "$P_s$", and the input pressure for the "control layer" is defined as "$P_c$". The source pressure $P_s$ for the normally-on valve is set to 100 Pa and for the normally-off valve is set to 10 Pa. Through parametric scanning, we obtain the membrane displacement at the top and the output flow rate $Q_o$ as functions of $P_c$ while keeping $P_s$ constant (Figure S1C and S1E). The output flow rate is determined by performing a "phase value - surface integral" calculation at the outlet, while the membrane displacement is obtained through "point evaluation".

The simulation trend of the "$Q_o$-$P_c$" relationship is generally consistent with Figure 2E and Figure 2F, where the curve can be clearly divided into three regions, analogous to the "cutoff," "linear," and "saturation" regions in Shockley's first-order transistor models. However, due to the mechanical interaction between control pressure and input pressure, some discrepancies remain.

The threshold pressure $P_t$ between the cutoff region and the linear region differs between the normally-on valve and the normally-off valve due to structure difference. For the normally-on valve, since the areas affected by the control pressure and input pressure on the membrane are similar when the valve is open, the threshold pressure required to close the valve $P_{pt}$ [kPa] should be higher than the input pressure $P_s$. Based on Figure S1C, we can derive the following equation:

$$P_{pt} = P_s + P_1 \tag{1}$$

From the simulation results (Figure S1C) and experiment results (Figure 2E) the range of $P_1$ is 0.5 kPa to 4 kPa.

For normally-off valve, since the area of membrane affected by source pressure $S_s$ is smaller than the area affected by control pressure $S_c$, a lower control pressure can be used to open the valve under high input pressure. The relationship between the threshold pressure $P_{nt}$ [kPa] required to open the valve and the input pressure is as follows:

$$P_{nt} = \frac{S_s}{S_c} P_s + P_2 \tag{2}$$

From the simulation results (Figure S1E) and experiment results (Figure 2F) the range of $P_2$ is 0.2 kPa to 4 kPa.

For the normally-on valve, the "linear" region exists only during the brief period when the membrane's protrusion gradually detaches from the upper wall of the input layer. This short duration causes the saturation threshold pressure $P_{psat}$ to be close to $P_{pt}$. Once the protrusion fully detaches from the upper wall, the valve immediately enters the "saturation" region. For the normally-off valve, when the valve opens, the membrane still needs to move further upward until reach the "saturation" region. From the results (Figure S1E), the saturation threshold pressure for the normally-off valve $P_{nsat}$ is calculated as follows:

$$P_{nsat} = P_s + P_3\ [kPa] \tag{3}$$

From the simulation results (Figure S1E) and experiment results (Figure 2F) the range of $P_2$ is 1 kPa to 5 kPa.

The equations describing the relationship between the output flow rate $Q_o$ and control pressure $P_c$ in each region are shown below.

For cutoff region:

$$Q_o = 0, \left(P_c \geq P_{pt}, \text{normally} - \text{on}\right) \tag{4}$$

$$Q_o = 0, \left(P_c < P_{pt}, \text{normally} - \text{off}\right) \tag{5}$$

For linear region:

$$Q_o = \beta_p\left(P_c - \alpha_p P_s\right)P_s, (P_{psat} \leq P_c < P_{pt}, \text{normally} - \text{on}) \quad (6)$$
$$Q_o = \beta_n(P_c - \alpha_n P_s)P_s, (P_{nt} < P_c \leq P_{nsat}, \text{normally} - \text{off}) \quad (7)$$

where $\beta_p$, $\beta_n$ are controlled by the structural parameters of the valves and $\alpha_p$, $\alpha_n$ are deter-mine by $S_s$ and $S_c$.

For saturation region:

$$Q_o = \frac{\beta_p}{2}(P_s - P_{pt}), (P_c < P_{psat}, \text{normally} - \text{on}) \quad (8)$$
$$Q_o = \frac{\beta_n}{2}(P_s - P_{nt}), (P_c > P_{nsat}, \text{normally} - \text{off}) \quad (9)$$

By adding the appropriate "point evaluation" to measure the pressure at specific nodes in the "input layer," we observe the corresponding changes in the flow resistance of the normally-on valve $R_{CP}$ and the normally-off valve $R_{CN}$ caused by the deformation of the membrane, as shown in Figure S1D and S1F.

**2. Comparison of the number of valves and inputs required for different basic logics designs.**

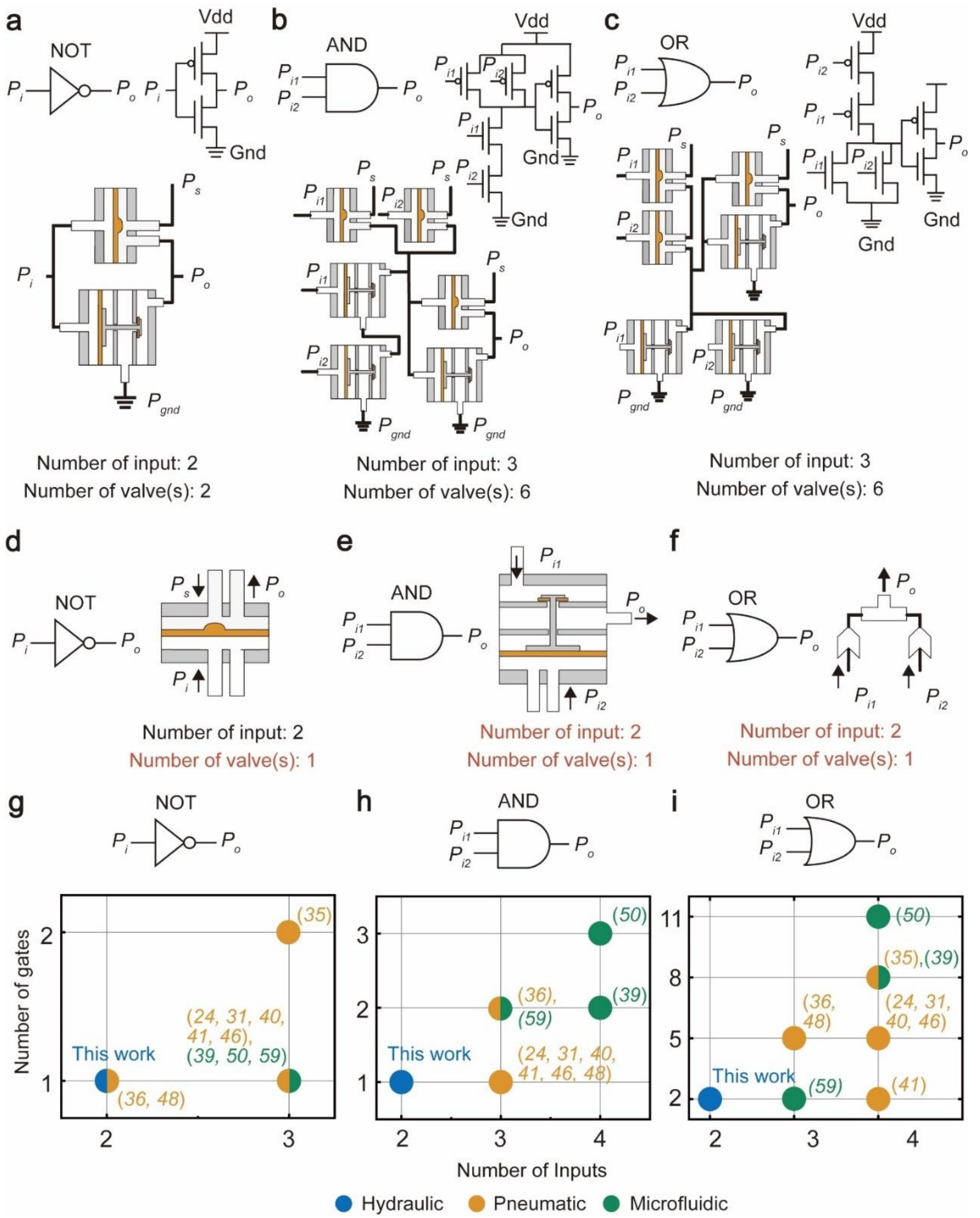

**Figure S2. Comparison of the number of valves and inputs required for different basic logics designs.** CMOS based (A) NOT gate, (B) AND gate, (C) OR gate, respectively.Single valve serves as (D) NOT gate, (E) AND gate, (F) OR gate, respectively. The comparison of the number of valves and inputs required for (G) NOT gate, (H) AND gate, (I) OR gate between this work and previous work. This work has the least number of inputs required for basic logic gates.

**3. Mechanism of gain generation by parallel active/passive flow amplifiers.**

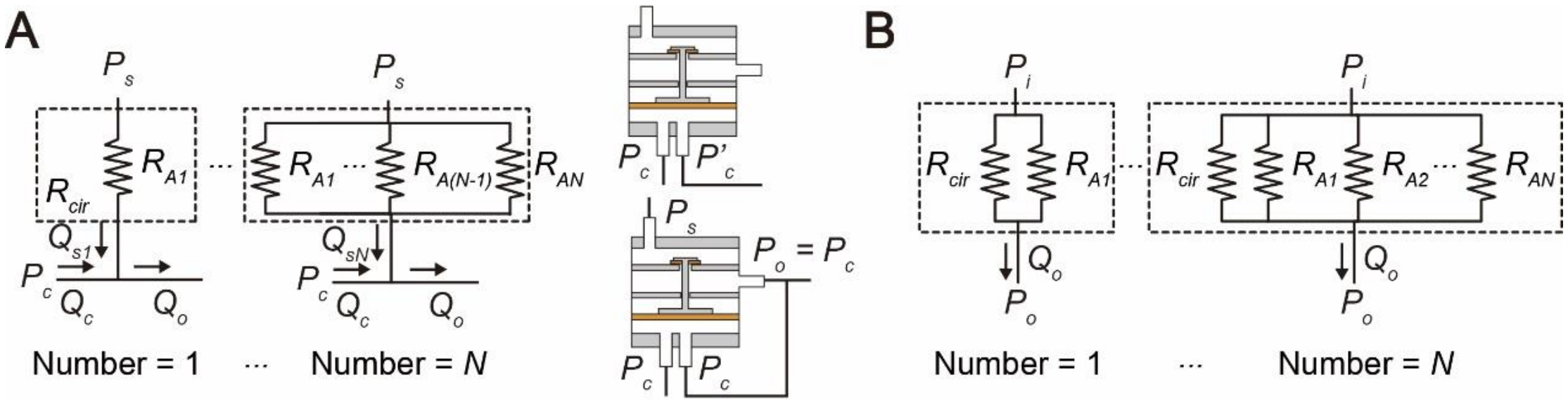

**Figure S3. Using equivalent flow resistance to explain the mechanism of gain generation by parallel active/passive flow amplifiers.** (A) The equivalent flow resistance analogy for gain generation by paralleling multiple active flow amplifiers. (B) The equivalent flow resistance analogy for gain generation in multiple parallel passive flow amplifiers.

We consider the active amplifier as a flow resistance within the hydraulic logic circuitry, with its equivalent circuit shown in Figure S3A. From Figure 7, B and C, the hydraulic pressures at each end of the amplifier are the source pressure $P_s$ and the output pressure $P_c$.

When only one active flow amplifier is used, the equivalent total flow resistance of the circuit $R_{cir}$ can be calculated as follows:

$$R_{cir1} = R_A \tag{10}$$

The amplified output flow rate can be calculated as follows:

$$Q_{o1} = Q_c + Q_{s1} = Q_c + \frac{P_s - P_c}{R_{cir}} \tag{11}$$

The active gain $A_{o1}$ can be calculated:

$$A_{o1} = \frac{Q_{o1}}{Q_c} = 1 + \frac{(P_s - P_c)}{R_{cir} \cdot Q_c} \tag{12}$$

Based on the theory of electrical circuits, when resistors are connected in parallel, the total resistance of the circuit is lower than any individual resistor. Similarly, in fluid circuits, connecting valves in parallel within the output path reduces the overall flow resistance.

For two or more active flow amplifiers are used, the equivalent total flow resistance of the circuit $R_{cir}$ can be calculated as follows:

$$R_{cirN} = \frac{R_A}{N} (N = 2,3, \dots, N) \tag{13}$$

The increase in flow gain when the $N^{th}$ ($N \geq 2$)valve is added is calculated as follows:

$$\Delta A_{oN} = \frac{Q_{oN}}{Q_c} - \frac{Q_{o(N-1)}}{Q_c} = \frac{(P_s - P_c)}{Q_c}\left(\frac{N}{R_A} - \frac{N-1}{R_A}\right) = \frac{P_s - P_c}{Q_c \cdot R_A} \tag{14}$$

The results in Figure 7D when the number of active amplifiers larger than two align with this conclusion. For the reason $\Delta A_{o1} > \Delta A_{o2}$, We believe that the active flow amplifier not only supplements the flow rate but also provides a certain gain in output pressure. This compensates for the pressure loss of $P_c$ as it passes through the amplifier's control layer, restoring the output pressure to $P_c$ and thereby increasing the original flow rate $Q_c$. This demonstrates that the flow gain introduced by a single parallel active flow amplifier remains constant and is determined by the initial input pressures $P_s$, $P_c$, initial output flowrate $Q_c$ and $R_A$ determined by valve structure and $P_c$ (Figure S1F). The gain is unaffected by the number of amplifiers in parallel, making it highly valuable for precise flow gain control.

For the equivalent resistance analogy of passive flow amplifiers, as shown in Figure S3B. After connecting *N* passive flow amplifiers in parallel, the equivalent total flow resistance of the circuit $R_{cirN}$:

$$R_{cirN} = \frac{R_{cir} \cdot R_A}{NR_{cir} + R_A} \tag{15}$$

The amplified output flow rate can be calculated as follows:

$$Q_{oN} = \frac{P_i - P_o}{R_{cirN}} = \frac{(P_i - P_o)(NR_{cir} + R_A)}{R_{cir} \cdot R_A} \tag{16}$$

The passive gain $A_{oN}$ can be calculated:

$$A_{oN} = \frac{Q_{oN}}{Q_{o0}} = \frac{NR_{cir} + R_A}{R_A} = 1 + \frac{NR_{cir}}{R_A} \tag{17}$$

It can be observed that a passive flow amplifier, even without additional input, can still increase the output flow rate. This effect depends on the flow resistance of the passive flow amplifier, the number of passive flow amplifiers, and the initial total flow resistance of the circuit.

4. System state analysis of hydraulic logic circuitry when driving a quadruped soft robot.

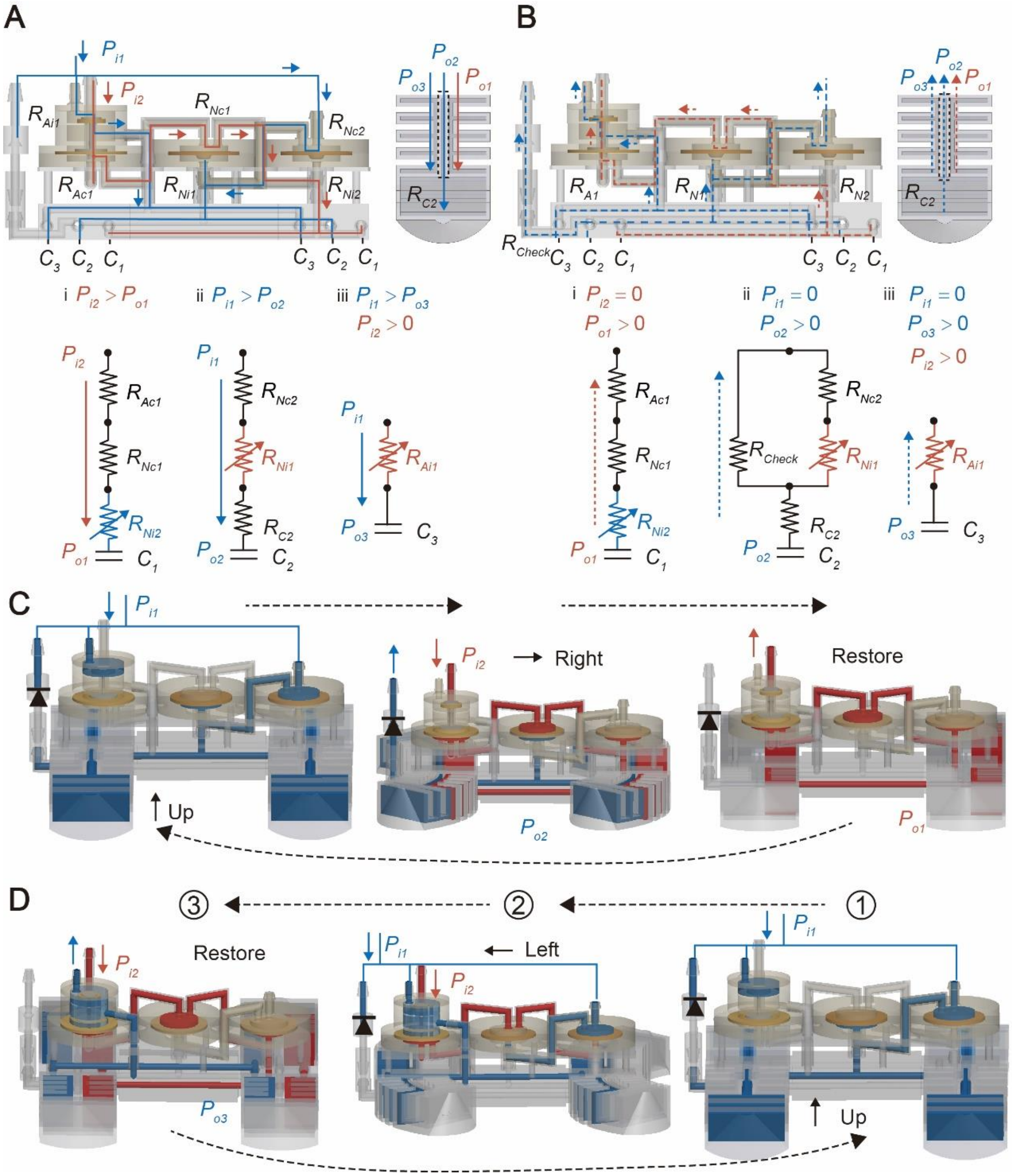

**Figure S4. System state analysis of hydraulic logic circuitry when driving a quadruped soft robot.** (A) Analysis of channel flow resistance when input pressure exceeds internal chamber pressure. (B) Analysis of channel flow resistance when input pressure is lower than internal chamber pressure. (C) Rightward alternating gait driven by input pressure and the rebound effect. (D) Rightward alternating gait driven by input pressure and the rebound effect.

In experiments using hydraulic logic circuitry to drive a quadruped soft robot, the input channels and corresponding equivalent flow resistances of valves for input pressures $P_{i1}$, $P_{i2}$ are shown in Figure S4A. The backflow channels and corresponding equivalent flow resistances of valves for output pressures $P_{o1}$, $P_{o2}$ and $P_{o3}$ are shown in Figure S4B. To prevent mutual interference in deformation, the actuator design separates the chamber controlling the robot's up-and-down movement from the chamber controlling its forward and backward movement. However, this approach introduces an additional segment of flow resistance in the input channel for chamber 2 $C_2$, prolonging the time required for the chamber to rebound. To maintain consistency in the deformation-rebound cycle across the three chambers, we add a check valve directly connected to the input end in parallel with the original return path for $P_{o2}$. This method reduces the flow resistance of the circuitry with minimal spatial impact on the system while preventing channel blockage caused by XOR operations when $P_{o2} > 0$ and $P_{i1}$ is applied.

**5. Comparison of the driving effects of hydraulic and pneumatic systems.**

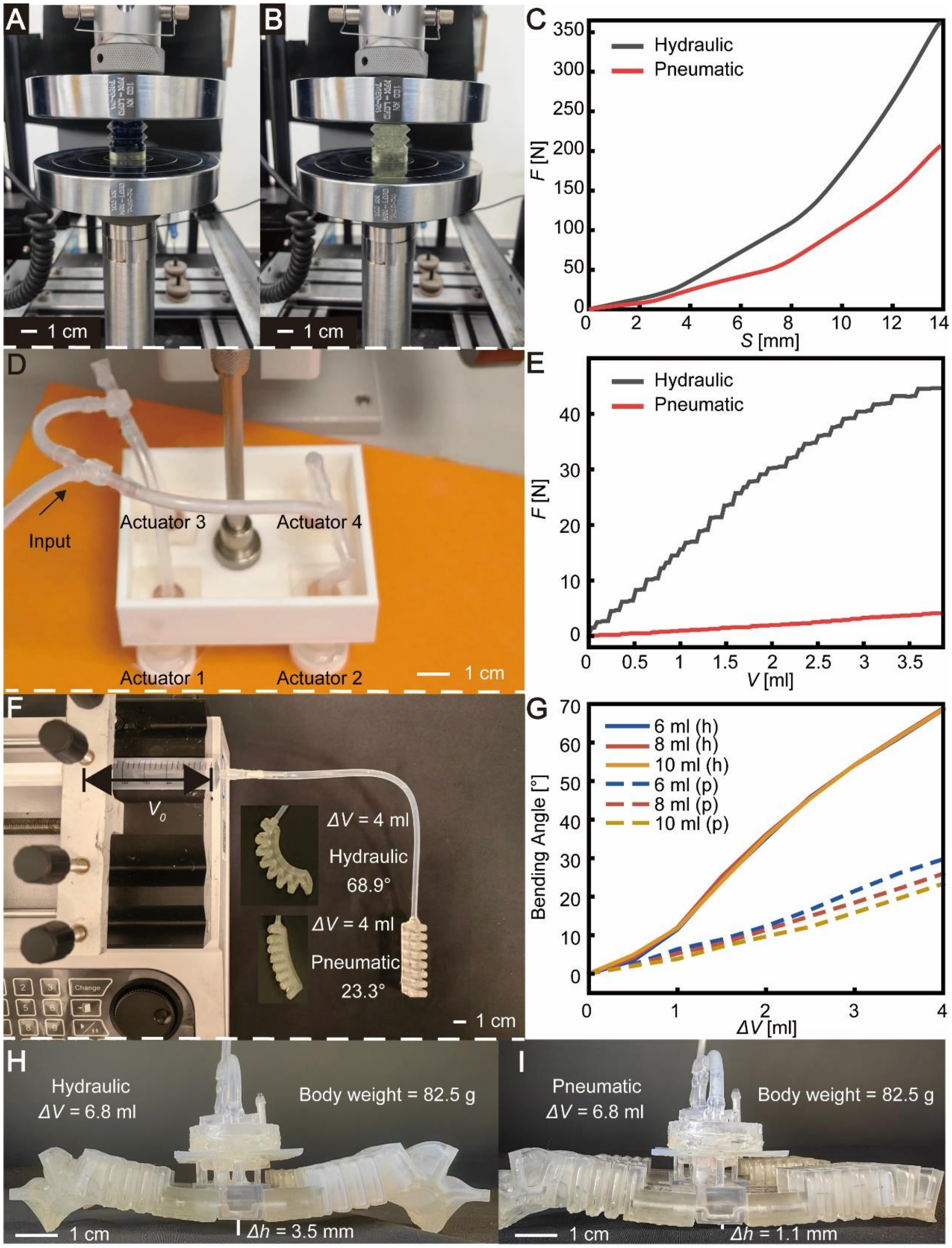

**Figure S5. Comparison of the driving effects of hydraulic and pneumatic systems.** (A) Using a compression testing machine to test the load that a hydraulic actuator can provide under zero input conditions. (B) Using a compression testing machine to test the load that a pneumatic actuator can provide under zero input conditions. (C) Comparison of Compression behavior of hydraulic actuators and pneumatic actuators under Load. (D) Comparing the support force generated on the platform by four actuators when injecting the same volume of water and air. (E) Results of the comparison between the support force

generated by hydraulic actuators and pneumatic actuators when injecting the same fluid. (F) Using a syringe pump to inject the same volume of liquid or gas into the actuators. (G) The comparing the bending angles of actuators injected with the same volume of gas and liquid under different initial conditions. Comparison the lifting heights of soft robots when injected with equal volumes of (H) water and (I) air.

**6. Experimental device.**

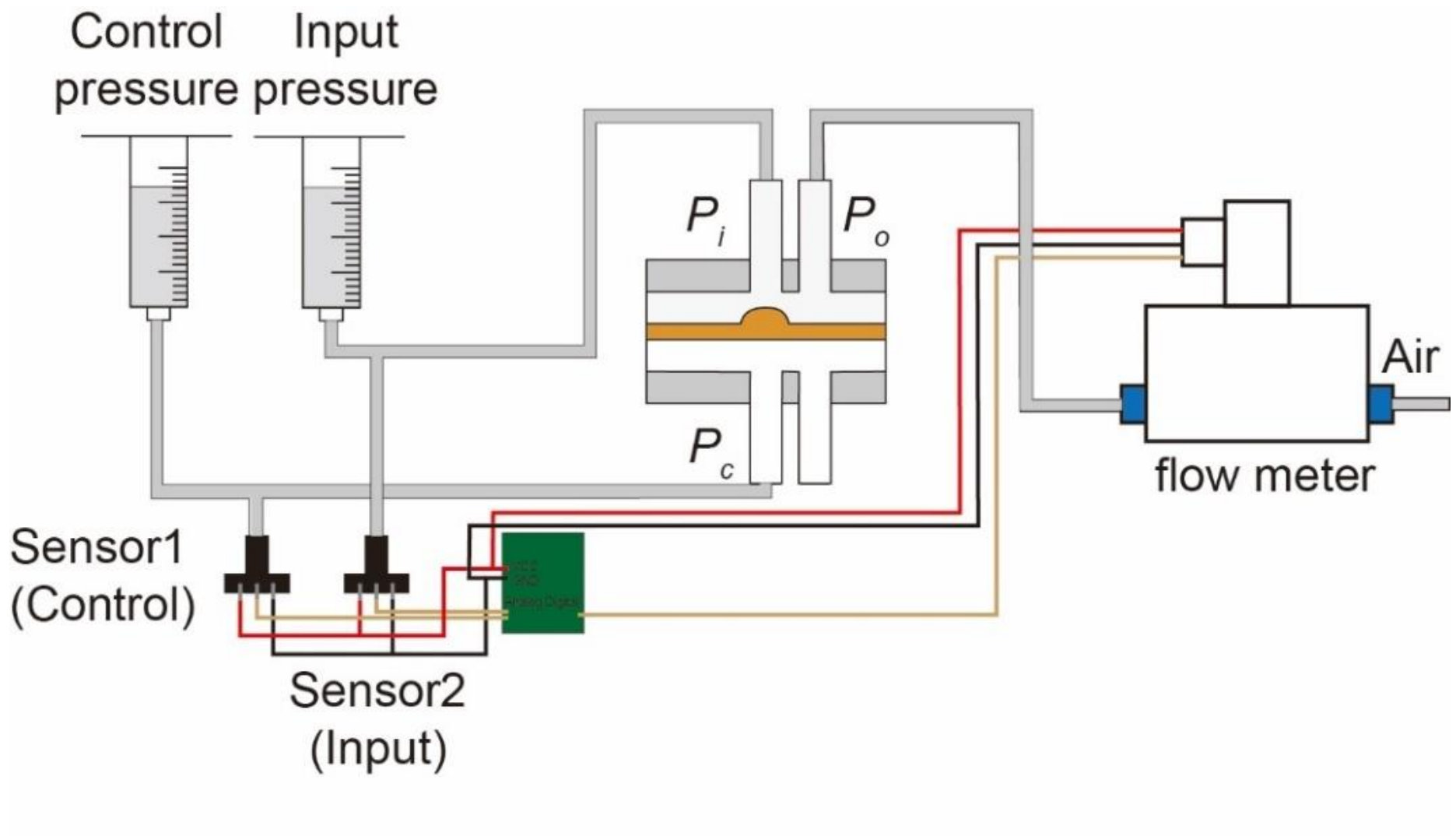

**Figure S6. Wiring diagram of the NOT gate testing.**

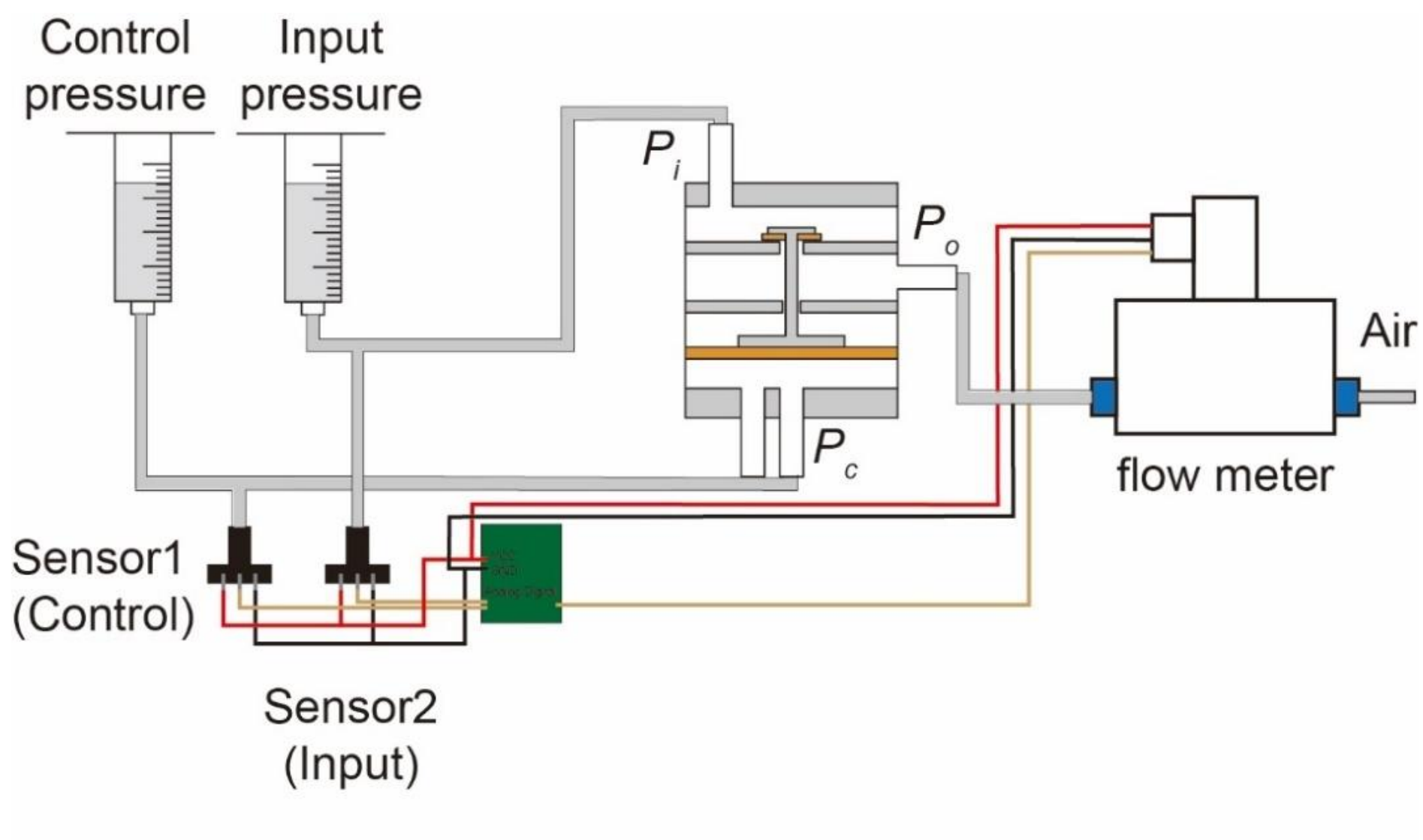

**Figure S7. Wiring diagram of the AND gate testing.**

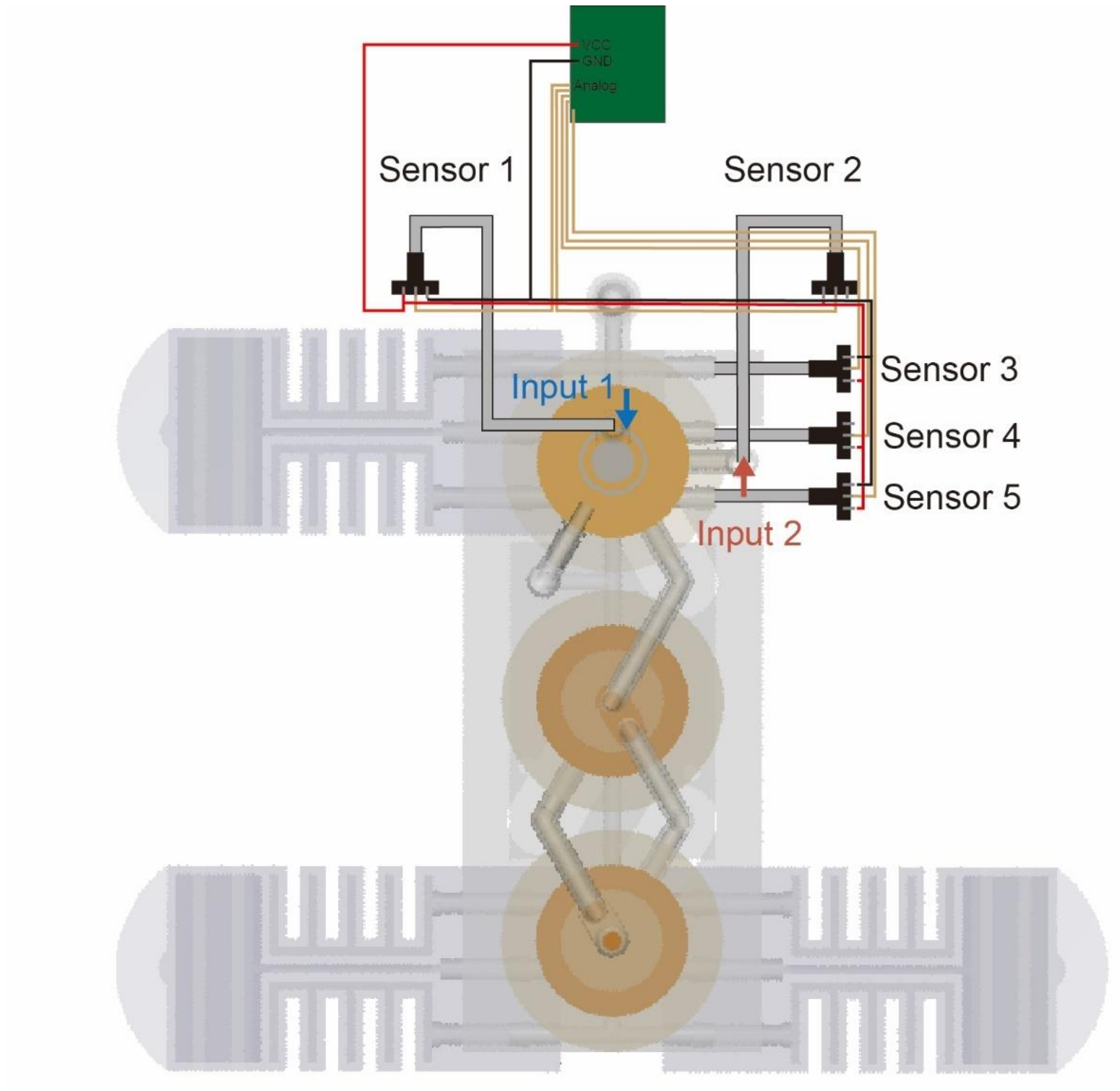

**Figure S8. Hydraulic logic circuitry for soft robot pressure testing system.**